%% file: main.tex
\documentclass[10pt,logo,copyright]{xhumanoid-report}
\input{packages}

\input{common.tex}
\definecolor{darkred}{rgb}{0.7, 0.0, 0.0}

\usepackage{caption}
\usepackage{amsmath} 
\usepackage{xcolor}
\usepackage{hyperref}
\definecolor{linkpink}{RGB}{255,90,170}
\usepackage{pifont}
\newcommand{\cmark}{\ding{51}} %
\newcommand{\xmark}{\ding{55}} %

\usepackage[nameinlink]{cleveref}
\crefname{equation}{Eq.}{Eqs.}
\crefname{figure}{Fig.}{Figs.}
\crefname{section}{Sec.}{Sec.}
\crefname{appendix}{App.}{App.}
\crefname{table}{Tab.}{Tabs.}
\crefname{algorithm}{Algo}{Algo}
\crefname{thm}{Thm}{Thm}
\Crefname{thm}{Thm}{Thm}
\crefname{prop}{Prop}{Prop}

\usepackage{placeins}

\newcommand{\crefnames}[3]{%
  \@for\next:=#1\do{%
    \expandafter\crefname\expandafter{\next}{#2}{#3}%
  }%
}

\title{Heracles: Bridging Precise Tracking and Generative Synthesis for General Humanoid Control}

\author{X-Humanoid Heracles Project Team}

\begin{abstract}

Achieving general-purpose humanoid control requires a delicate balance between the precise execution of commanded motions and the flexible, anthropomorphic adaptability needed to recover from unpredictable environmental perturbations. Current general controllers predominantly formulate motion control as a rigid reference-tracking problem. While effective in nominal conditions, these trackers often exhibit brittle, non-anthropomorphic failure modes under severe disturbances, lacking the generative adaptability inherent to human motor control. To overcome this limitation, we propose Heracles, a novel state-conditioned diffusion middleware that bridges precise motion tracking and generative synthesis. Rather than relying on rigid tracking paradigms or complex explicit mode-switching, Heracles operates as an intermediary layer between high-level reference motions and low-level physics trackers. By conditioning on the robot's real-time state, the diffusion model implicitly adapts its behavior: it approximates an identity map when the state closely aligns with the reference, preserving zero-shot tracking fidelity. Conversely, when encountering significant state deviations, it seamlessly transitions into a generative synthesizer to produce natural, anthropomorphic recovery trajectories. Our framework demonstrates that integrating generative priors into the control loop not only significantly enhances robustness against extreme perturbations but also elevates humanoid control from a rigid tracking paradigm to an open-ended, generative general-purpose architecture.

\end{abstract}

\begin{document}

\maketitle

\abscontent
\input{sections/1_introduction}
\input{sections/2_related_work}
\input{sections/3_methodology}
\input{sections/4_experiments_and_results}
\input{sections/5_conclusion}
\input{sections/6_project_team}
\input{sections/6_ack}

\clearpage



\clearpage
\setcitestyle{numbers}
\bibliographystyle{plainnat}
\bibliography{main}

\end{document}

%% file: packages.tex
\usepackage[authoryear,sort&compress,round]{natbib}

\usepackage[utf8]{inputenc} %
\usepackage[T1]{fontenc}    %

\usepackage{parskip}        %
\usepackage{url}            %
\usepackage{booktabs}       %
\usepackage{amsfonts}       %
\usepackage{nicefrac}       %
\usepackage{microtype}      %
\usepackage{xcolor}         %
\usepackage[dvipsnames]{xcolor} %
\usepackage{graphicx}
\usepackage{animate}        %
\usepackage{subcaption}
\usepackage{tabularx}
\usepackage{makecell}
\usepackage{adjustbox}
\usepackage{setspace}
\usepackage{todonotes}
\usepackage{colortbl}
\usepackage{wrapfig}

\newcolumntype{M}[1]{>{\centering\arraybackslash}m{#1}}
\usepackage{float}
\usepackage{tikz}
\usetikzlibrary{positioning,shapes,arrows}
\usepackage{amsmath,amsfonts,bm, bbm,leftindex}
\usepackage{multirow}
\usepackage{comment}
\usepackage{gensymb}
\usepackage{lipsum}
\usetikzlibrary{arrows.meta, positioning, fit}
\usepackage[para]{threeparttable}
\usepackage{tikz}
\usetikzlibrary{tikzmark}

%% file: common.tex
\def\eqref#1{equation~\ref{#1}}

\def\1{\bm{1}}

\DeclareMathAlphabet{\mathsfit}{\encodingdefault}{\sfdefault}{m}{sl}
\SetMathAlphabet{\mathsfit}{bold}{\encodingdefault}{\sfdefault}{bx}{n}

\makeatletter
\let\save@mathaccent\mathaccent
\newcommand*\if@single[3]{%
  \setbox0\hbox{${\mathaccent"0362{#1}}^H$}%
  \setbox2\hbox{${\mathaccent"0362{\kern0pt#1}}^H$}%
  \ifdim\ht0=\ht2 #3\else #2\fi
  }
\newcommand*\rel@kern[1]{\kern#1\dimexpr\macc@kerna}
\newcommand*\widebar[1]{\@ifnextchar^{{\wide@bar{#1}{0}}}{\wide@bar{#1}{1}}}
\newcommand*\wide@bar[2]{\if@single{#1}{\wide@bar@{#1}{#2}{1}}{\wide@bar@{#1}{#2}{2}}}
\newcommand*\wide@bar@[3]{%
  \begingroup
  \def\mathaccent##1##2{%
    \let\mathaccent\save@mathaccent
    \if#32 \let\macc@nucleus\first@char \fi
    \setbox\z@\hbox{$\macc@style{\macc@nucleus}_{}$}%
    \setbox\tw@\hbox{$\macc@style{\macc@nucleus}{}_{}$}%
    \dimen@\wd\tw@
    \advance\dimen@-\wd\z@
    \divide\dimen@ 3
    \@tempdima\wd\tw@
    \advance\@tempdima-\scriptspace
    \divide\@tempdima 10
    \advance\dimen@-\@tempdima
    \ifdim\dimen@>\z@ \dimen@0pt\fi
    \rel@kern{0.6}\kern-\dimen@
    \if#31
      \overline{\rel@kern{-0.6}\kern\dimen@\macc@nucleus\rel@kern{0.4}\kern\dimen@}%
      \advance\dimen@0.4\dimexpr\macc@kerna
      \let\final@kern#2%
      \ifdim\dimen@<\z@ \let\final@kern1\fi
      \if\final@kern1 \kern-\dimen@\fi
    \else
      \overline{\rel@kern{-0.6}\kern\dimen@#1}%
    \fi
  }%
  \macc@depth\@ne
  \let\math@bgroup\@empty \let\math@egroup\macc@set@skewchar
  \mathsurround\z@ \frozen@everymath{\mathgroup\macc@group\relax}%
  \macc@set@skewchar\relax
  \let\mathaccentV\macc@nested@a
  \if#31
    \macc@nested@a\relax111{#1}%
  \else
    \def\gobble@till@marker##1\endmarker{}%
    \futurelet\first@char\gobble@till@marker#1\endmarker
    \ifcat\noexpand\first@char A\else
      \def\first@char{}%
    \fi
    \macc@nested@a\relax111{\first@char}%
  \fi
  \endgroup
}
\makeatother

%% file: sections/1_introduction.tex
\section{Introduction}

Humanoid robots are rapidly transitioning from structured laboratory settings to complex, unstructured real-world environments. This shift demands general-purpose control architectures capable of executing precise, goal-directed motions while maintaining the flexible, anthropomorphic resilience characteristic of human motor control. In biological systems, motor behavior is rarely a rigid execution of a predefined plan; rather, humans seamlessly blend exact task execution with intuitive, generative recovery when faced with unexpected physical disturbances. Emulating this dual capability remains a formidable challenge in modern robotics.

The prevailing paradigm for general-purpose humanoid control relies heavily on reference-driven tracking. Recent advancements formulate motion control primarily as a problem of minimizing the kinematic deviation between the robot's current state and a provided reference trajectory. Powered by deep reinforcement learning, these tracking-based controllers~\cite{yin2025unitracker,luo2025sonic,wang2026omnixtreme,he2024hover}, such as those mimicking reference motions or utilizing universal trackers, excel in nominal conditions. They enable humanoids to faithfully replicate highly diverse sets of motion capture (MoCap) data, achieving impressive zero-shot execution for an array of highly agile and dynamic skills.

\begin{figure*}[!htbp]
\centering
\includegraphics[width=0.8\linewidth]{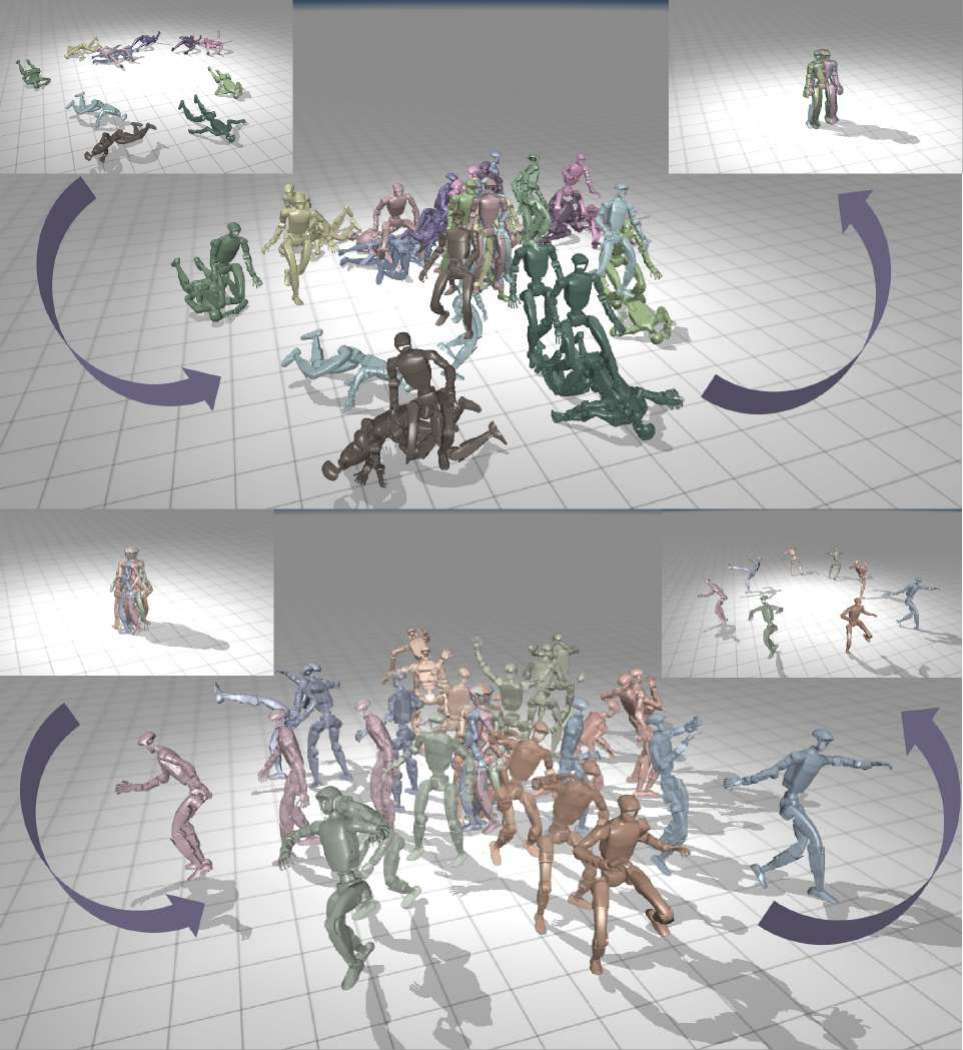}
\captionsetup{justification=raggedright,singlelinecheck=false}
\caption{\small \textbf{Heracles synthesizes diverse, anthropomorphic recovery motions via state-conditioned diffusion.} In contrast to recovery policies trained with termination tricks that converge to a limited set of stereotypical maneuvers, Heracles leverages its generative middleware to produce a rich repertoire of agile, human-like recovery behaviors, enabling more general and robust responses across a wide range of extreme perturbations.}
\label{fig:teaser}
\end{figure*}

However, formulating general control strictly as a rigid tracking objective introduces a critical vulnerability: catastrophic and non-anthropomorphic failure modes under severe environmental perturbations. When a robot is pushed far from its reference trajectory, a pure tracker blindly attempts to minimize the immediate state error, often resulting in rigid, physically infeasible joint torques that lead to unnatural and unrecoverable falls. Conversely, entirely dropping the tracking objective in favor of pure generative models~\cite{zeng2025bfm,li2025bfmzero}, such as end-to-end behavior cloning, yields more natural, human-like movements but fundamentally sacrifices the spatial and temporal precision required for strict task execution. A significant gap remains: bridging the exactitude of precise trackers with the generative adaptability of robust human recovery.

To overcome this fundamental limitation, we propose Heracles, a novel state-conditioned diffusion middleware designed to bridge precise motion tracking and generative synthesis (\cref{fig:teaser}). Rather than engineering a monolithic end-to-end controller or relying on complex, explicit state machines to switch between tracking and recovery modes, Heracles operates elegantly as an intermediary layer. Situated between the high-level original motion commands and the low-level physical execution policy, it injects powerful generative priors directly into the control loop without disrupting the high-frequency physics execution.

The core innovation of Heracles lies in its implicit, state-driven adaptability. By tightly conditioning the diffusion process on the robot's real-time physical state, the model dynamically shapes its output without explicit heuristics. When the robot's state closely aligns with the reference command, the diffusion process approximates an identity map, allowing the commands to pass through with near-zero modification to preserve strict tracking fidelity. Conversely, when encountering significant state deviations—such as a severe push or an imminent fall—the model seamlessly transitions into a generative synthesizer. It synthesizes natural, anthropomorphic recovery trajectories that guide the underlying physics tracker back to stability. This mechanism not only enhances physical robustness but elevates humanoid control from a rigid tracking paradigm to an open-ended, generative architecture.

In summary, the primary contributions of this work are three-fold:

\begin{itemize}
    \item \textbf{A Generative Control Middleware Paradigm:} We introduce Heracles, a novel state-conditioned diffusion middleware that uniquely bridges the precision of motion tracking with the flexibility of generative synthesis, effectively decoupling high-level intent generation from low-level physical execution.

    \item \textbf{Enhanced Architecture for General-Purpose Control:} We improve the underlying physics tracker and the overall control framework to better serve general-purpose tasks. This optimized architecture ensures the retention of high-fidelity tracking characteristics while seamlessly integrating with the generative priors from the middleware.

    \item \textbf{Robust Anthropomorphic Recovery and Motion Generalization:} We successfully deploy the proposed framework on physical humanoid robots. Extensive hardware experiments demonstrate emergent, human-like recovery behaviors and robust generative adaptability under severe, out-of-distribution physical disturbances.
\end{itemize}

%% file: sections/2_related_work.tex
\section{Related Work}
\subsection{General Humanoid Motion Controller}
Recent advances in deep reinforcement learning catalyze the development of general-purpose controllers for high-degree-of-freedom humanoid robots. These architectures aim to provide a unified execution policy for diverse motor behaviors, predominantly dividing into mimic-based reference trackers and unsupervised reinforcement learning (URL) methods.

\textbf{Mimic-Based Motion Trackers.} A dominant paradigm formulates motor behavior execution as a high-fidelity motion imitation task. Pioneered by DeepMimic~\cite{peng2018deepmimic}, which demonstrated that deep reinforcement learning can train physics-based characters to closely imitate reference motion clips, this paradigm has since been scaled to general-purpose humanoid control. Foundational large-scale frameworks, including the Generalized Motion Tracker (GMT)~\cite{chen2025gmt}, demonstrate that adaptive sampling and mixture-of-experts (MoE) architectures enable humanoids to track diverse motions via a single unified policy. Extending this paradigm, recent efforts structurally scale these models. SONIC~\cite{luo2025sonic} expands network capacity and datasets to over 100 million frames, establishing motion tracking as a robust scalable foundation task. To break the representation bottleneck in multi-motion RL optimization, OmniXtreme~\cite{wang2026omnixtreme} introduces a flow-matching policy to decouple general motor skill learning from sim-to-real physical refinement. Beyond pure kinematic tracking, researchers continuously enhance the versatility and stability of these frameworks. HOVER~\cite{he2024hover} employs multi-mode policy distillation to consolidate specific control modes—specifically navigation and loco-manipulation—into a unified controller. Concurrently, AMS~\cite{pan2025ams} proposes a hybrid reward scheme combining agile human MoCap data with physically constrained synthetic balance motions. More recently, BeyondMimic~\cite{liao2025beyondmimic} integrates a guided diffusion mechanism directly into the tracking formulation, allowing the system to solve diverse downstream tasks via classifier guidance. Beyond pure kinematics, ~\cite{sun2024prompt} pioneer the integration of large language models with adversarial imitation learning for zero-shot task execution through quantized skill representations. RoboGhost~\cite{li2025language} integrates language-grounded motion latents from a motion generator with reinforcement learning for  retarget-free whole-body controller. OmniRetarget~\cite{omniretarget} incorporates human-scene interaction (HSI) and human-object interaction (HOI) constraints into the motion retargeting pipeline, improving the physical plausibility of transferred motions. Along similar lines, recent works have further advanced humanoid parkour-style locomotion~\cite{zhuang2026deep, zhu2026ttt} and contact-rich interaction tasks~\cite{lin2026lessmimic}. MeshMimic~\cite{zhang2026meshmimic} further advances the paradigm by incorporating 3D scene reconstruction from monocular video, enabling humanoid robots to learn coupled motion-terrain interactions on complex, non-flat terrains. Despite these structural advancements, tracking-first controllers remain fundamentally anchored to their reference trajectories and typically require computationally expensive test-time guidance for generation. Under severe out-of-distribution physical perturbations, relying purely on the tracking formulation predominantly results in rigid, physically infeasible joint torques rather than seamless, real-time anthropomorphic recovery.

\textbf{Unsupervised RL and Skill Discovery.} Conversely, unsupervised reinforcement learning (URL) and the emerging paradigm of Behavioral Foundation Models (BFMs) seek to equip robots with a diverse repertoire of motor skills devoid of explicit reference trajectories. Recent extensive surveys on BFMs~\cite{yuan2025bfmsurvey} delineate a clear trajectory toward leveraging large-scale pre-training to capture broad behavioral priors. ~\cite{zhang2024wholebody} extended this paradigm to full-size humanoid robots by developing an adversarial motion prior framework that achieves human-comparable whole-body locomotion performance. The foundational BFM framework~\cite{zeng2025bfm} utilizes masked online distillation alongside Conditional Variational Autoencoders (CVAEs) to model behavioral distributions flexibly from unstructured data. Pushing the boundaries of autonomous exploration, cutting-edge URL approaches establish entirely reference-free policies. BFM-Zero~\cite{li2025bfmzero} leverages unsupervised RL and Forward-Backward (FB) representations to create an objective-centric, promptable latent space, enabling a single generalist policy to perform zero-shot tasks and reward inference seamlessly in the real world. Several recent works~\cite{luo2024smplolympics,chen2026learning,wang2025skillmimic,yu2025skillmimic,zhang2026learning} leverage motion tracking as a foundational mechanism to acquire human athletic skills, enabling humanoid robots to perform highly dynamic ball sports. Because these URL-driven policies and foundation models explore the state-action space unconstrained by rigid references, they inherently exhibit remarkable physical compliance and naturalistic robustness when perturbed. However, mapping these autonomously discovered, unconstrained latent spaces to high-precision, strict-fidelity spatial tracking tasks remains a formidable challenge. They fundamentally struggle to achieve the exactitude characteristic of dedicated mimic controllers in complex, dynamic execution scenarios.

\subsection{Motion Generation}
Synthesizing diverse, naturalistic human movements represents a foundational pursuit within computer animation. Motion-X~\cite{lin2023motion} provides a large-scale multi-modal human motion dataset comprising over 81K motion sequences with unified whole-body annotations spanning face, hands, and body, while its successor Motion-X++~\cite{zhang2025motion} further extends the scale and diversity by incorporating additional motion sources and richer semantic labels to support more comprehensive whole-body motion generation and understanding. Early paradigm shifts leveraged score-based generative models, with the Human Motion Diffusion Model (MDM)~\cite{tevet2023mdm} establishing a robust transformer-based baseline for text-driven kinematic synthesis. Subsequently, researchers integrated motion generation into the Large Language Model (LLM) ecosystem. MotionGPT~\cite{jiang2023motiongpt} first demonstrated that treating human motion as a foreign language and unifying motion-text tasks via discrete tokenization yields strong multi-task performance. Building upon this insight, MotionGPT-2~\cite{wang2024motiongpt2} further quantizes multimodal inputs—including text and single-frame poses—into LLM-interpretable tokens for unified generation and understanding. Meanwhile, the MotionMillion framework~\cite{fan2025motionmillion} demonstrates that million-scale high-quality datasets coupled with autoregressive architectures unlock unprecedented zero-shot capabilities. Expanding modality fusion, OmniMotion~\cite{li2025omnimotion} utilizes a continuous masked autoregressive transformer to seamlessly integrate text, speech, and music into a cohesive whole-body generation mechanism. Pushing model capacity limits, HY-Motion 1.0~\cite{wen2025hymotion} successfully scales diffusion transformer-based flow matching models to the billion-parameter regime, yielding instruction-following digital animations with unparalleled fidelity. Despite achieving remarkable anthropomorphism and diversity, these generative foundation models fundamentally operate in an open-loop, pure kinematic domain. They synthesize trajectories comprising joint angles and global translations completely devoid of physical constraints. Implementing these raw kinematic outputs directly on a physical humanoid invariably triggers dynamics mismatches, leading to instability and falls, because the generative process ignores crucial embodied parameters including joint torque limits, contact friction, and center-of-mass dynamics.

\subsection{Hybrid Architectures for Motion Tracking and Synthesis}
To overcome the inherent limitations of isolated physical trackers and open-loop generative models, recent research actively constructs hybrid architectures integrating generative synthesis with tracking objectives. Within the kinematic domain, frameworks attempt to constrain generation via tracking formulations; COMET~\cite{lee2025comet} employs a conditional VAE framework with a reference-guided feedback loop to prevent long-term motion degradation, while MotionStreamer~\cite{xiao2025motionstreamer} and DART~\cite{zhao2024dart} enforce sequential synthesis driven by rigorous spatial constraints. Transitioning to physically simulated characters, building upon the foundational reference-tracking paradigm of DeepMimic~\cite{peng2018deepmimic}, Adversarial Skill Embeddings (ASE)~\cite{peng2022ase} constructs continuous latent spaces from large-scale unstructured motion data, enabling characters to maintain highly anthropomorphic behaviors during diverse downstream tasks. Advancing this trajectory, the Versatile Motion Priors (VMP) framework~\cite{serifi2024vmp} optimizes the robust control of physical characters by distilling multipurpose motion priors through a two-stage variational approach, significantly enhancing both reference trajectory tracking and resilience against external perturbations. Recent investigations further deepen this paradigm: AMOR~\cite{alegre2025amor} proposes multi-objective reinforcement learning to train weight-conditioned policies spanning the Pareto front of reward trade-offs, while adversarial differential discriminators~\cite{zhang2025add} eliminate the need for manually-designed reward functions in physics-based motion imitation. In the multi-agent competitive domain, RoboStriker~\cite{yin2026robostriker} constructs a hierarchical framework that decouples high-level strategic reasoning from low-level physical execution via topologically constrained latent manifolds, demonstrating emergent boxing behaviors with sim-to-real transfer. However, these hybrid control paradigms retain structural deficiencies when confronting extreme, out-of-distribution physical disturbances. They typically couple high-level generative priors with low-level execution policies loosely, preventing high-frequency physical state deviations from reshaping the generative target in real time. Consequently, when encountering severe imbalance, the system fails to transition implicitly from nominal tracking to generative recovery, often reverting to rigid, explicit state-switching mechanisms. Our proposed Heracles framework resolves this bottleneck through a state-conditioned diffusion middleware that dynamically modulates the generative output based on real-time state deviations, achieving a seamless unification of precise zero-shot tracking and anthropomorphic generative recovery within a closed control loop.

%% file: sections/3_methodology.tex
\section{Method}

\begin{figure*}[!htbp]
\centering
\includegraphics[width=\linewidth]{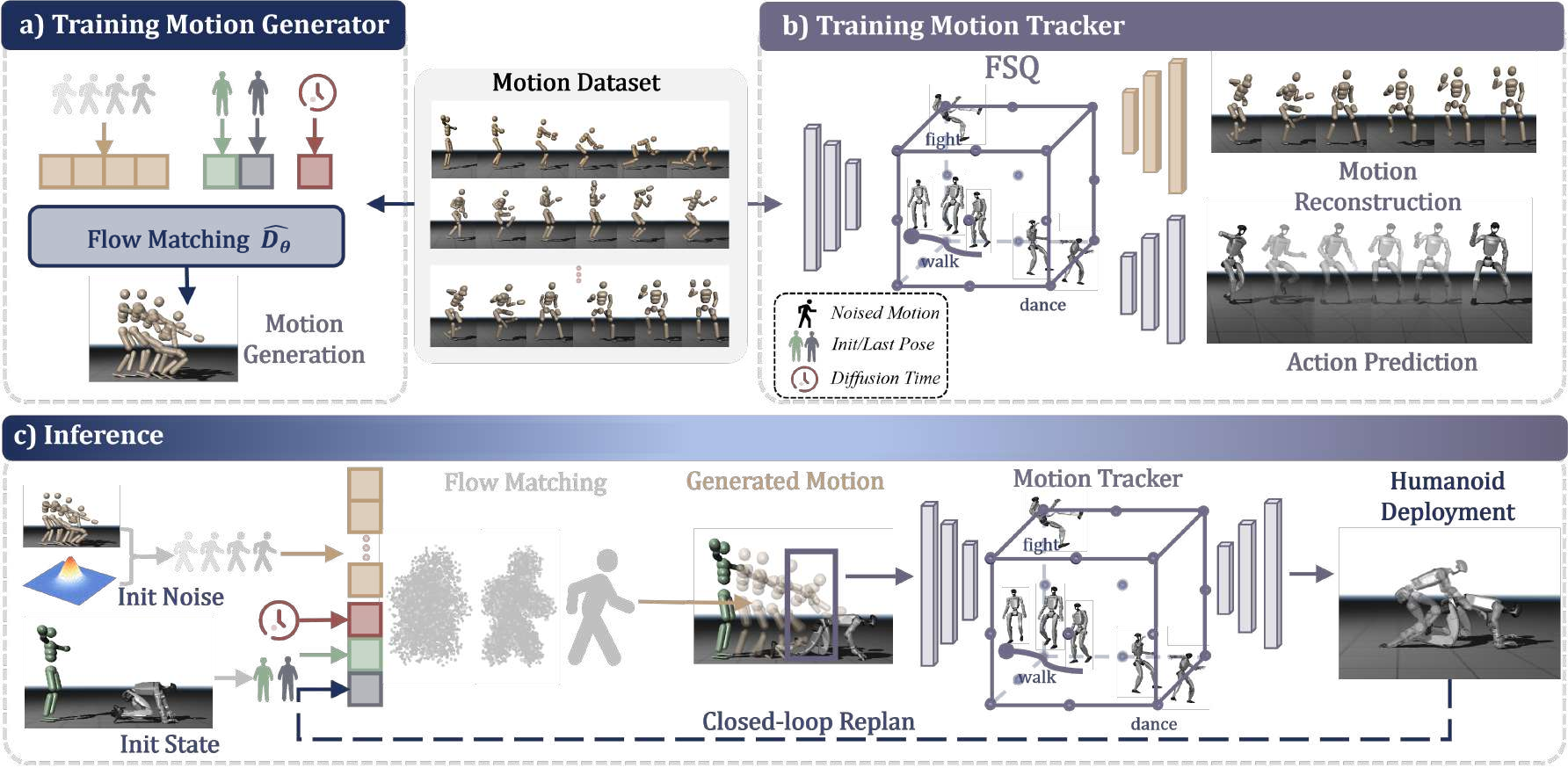}
\captionsetup{justification=raggedright,singlelinecheck=false}
\caption{\small \textbf{Overview of the Heracles framework.}
\textbf{(a)}~A flow matching model $\hat{D}_\theta$ learns to synthesize feasible keyframe trajectories conditioned on the current state.
\textbf{(b)}~Reference motions are quantized into discrete tokens via FSQ, shared by reconstruction and action prediction heads.
\textbf{(c)}~At inference, the middleware generates trajectories through closed-loop replanning for the motion tracker to execute.}
\label{fig:method_pipeline}
\end{figure*}

\subsection{System Overview and Problem Formulation}

General humanoid control requires executing desired motion commands while maintaining physical balance against unpredictable environmental disturbances. We formulate this dual objective—high-fidelity motion tracking and robust physical recovery—within a hierarchical control architecture. The proposed framework, Heracles, intrinsically decouples high-level intent generation from high-frequency physical execution (\cref{fig:method_pipeline}). It comprises two primary components: a state-conditioned generative middleware and a low-level, general-purpose physics tracking policy.

In standard tracking paradigms~\cite{peng2018deepmimic,chen2025gmt,luo2025sonic}, a policy minimizes the kinematic deviation between the robot's real-time proprioceptive state $\mathbf{p}_t$ and a reference motion command $\mathbf{m}_t$. When the state remains close to the reference manifold, directly tracking $\mathbf{m}_t$ yields optimal zero-shot performance. However, when severe physical perturbations push the state into out-of-distribution (OOD) regions, forcing strict adherence to $\mathbf{m}_t$ produces rigid, myopic corrective actions that lack the coordinated multi-step reasoning required for physical recovery, invariably precipitating catastrophic failure.

To resolve this limitation, we formulate the core problem as learning an intermediary mapping that dynamically modulates the reference command based on real-time physical feasibility. The generative middleware functions as this state-conditioned trajectory synthesizer. Operating at a lower planning frequency, it observes the current state $\mathbf{p}_t$ and the original reference $\mathbf{m}_t$, predicting a short-horizon, dynamically feasible keyframe trajectory:
\begin{equation}
\boldsymbol{\tau}_t = f_{\theta}(\mathbf{p}_t, \mathbf{m}_t) \in \mathbb{R}^{K \times D},
\label{eq:method_middleware}
\end{equation}
where $K$ denotes the number of keyframes and $D$ represents the state dimension. This mapping design unifies nominal tracking and OOD recovery: when the robot operates near the reference manifold, $f_{\theta}$ approximates an identity transformation to preserve high-fidelity tracking; conversely, under large disturbances, $f_{\theta}$ synthesizes entirely new, physically feasible transition trajectories that guide the robot back toward the reference manifold.

The synthesized keyframes $\boldsymbol{\tau}_t$ are subsequently densified and written into a reference buffer consumed by the low-level physics tracker. We model this continuous tracking process as a discounted Markov decision process (MDP) defined by the tuple $\mathcal{M}=(\mathcal{S},\mathcal{A},\mathcal{P},r,\gamma)$. At each high-frequency control step, the tracking policy $\pi$ receives an observation:
\begin{equation}
\mathbf{o}_t = \{\mathbf{p}_t,\, \mathbf{m}'_t,\, \mathbf{z}_d\},
\label{eq:method_obs}
\end{equation}
where $\mathbf{m}'_t$ denotes the modulated reference commands sampled from the densified $\boldsymbol{\tau}_t$, and $\mathbf{z}_d$ constitutes a high-level motion embedding (detailed in \cref{sec:tracker}). The policy outputs joint-level actions $a_t \in \mathcal{A}$ to maximize the expected discounted return:
\begin{equation}
J(\pi)=\mathbb{E}_{\tau\sim\pi}\!\left[\sum_{t=0}^{T-1}\gamma^t r_t\right],
\label{eq:method_rl_obj}
\end{equation}
where $r_t$ represents a task reward composed of tracking precision and physical regularization terms. Deployment follows a receding-horizon replanning loop. Every $N_{\mathrm{exec}}$ control steps, the middleware updates $\boldsymbol{\tau}_t$ from the latest proprioceptive observation, while the tracking policy executes the dense trajectory at the fundamental control frequency, forming a seamless, closed-loop tracking-generation architecture.

\subsection{State-Conditioned Generative Middleware via Flow Matching}

We formulate the trajectory generation process as a conditional flow matching problem over a geometrically constrained residual space, bridging exact motion tracking and generative synthesis without relying on explicit mode-switching heuristics.

\textbf{Geometric Residual Parameterization.}
Directly predicting absolute state coordinates wastes model capacity on approximating the identity mapping during near-nominal execution. Instead, we predict residual trajectories relative to the current state. Let $\boldsymbol{\beta}_{t}$ denote the static baseline anchored at the current proprioceptive state:
\begin{equation}
\boldsymbol{\beta}_{t,k} = \mathbf{p}_t, \qquad \forall\, k \in \{0, \ldots, K{-}1\}.
\label{eq:method_beta}
\end{equation}
The middleware predicts only the residual deviation $\mathbf{r}_t$, recovering the final trajectory as:
\begin{equation}
\boldsymbol{\tau}_t = \boldsymbol{\beta}_t + \mathbf{r}_t.
\label{eq:method_residual}
\end{equation}
Under this parameterization, the residual directly encodes the motion increment from the current state over the planning horizon. When the robot closely follows the reference manifold ($\mathbf{p}_t \approx \mathbf{m}_t$), the synthesized trajectory closely reproduces the original reference motion, and the middleware effectively acts as an identity map on the command signal. Under severe deviations, the model leverages the conditioning gap between $\mathbf{p}_t$ and $\mathbf{m}_t$ to synthesize recovery trajectories that diverge from the reference in favor of physical plausibility. Crucially, the target command $\mathbf{m}_t$ enters exclusively through the conditioning vector, ensuring that the residual prediction target remains independent of the state-command distance.

\textbf{Continuous Conditional Flow Matching.}
To synthesize the complex, multimodal distributions of these recovery residuals, we employ continuous flow matching~\cite{lipman2023flow}. Let $\mathbf{x}_0$ denote the normalized ground-truth residual data and $\mathbf{x}_1 \sim \mathcal{N}(0,\mathbf{I})$ represent the prior Gaussian noise. We define a probability path via linear interpolation:
\begin{equation}
\mathbf{x}_t = (1-t)\mathbf{x}_0 + t\mathbf{x}_1, \qquad t\in[0,1].
\label{eq:method_flow_path}
\end{equation}
The model is trained to regress the underlying continuous vector field by minimizing the velocity matching objective:
\begin{equation}
\mathcal{L}_{\mathrm{vel}} = \mathbb{E}_{t,\mathbf{x}_0,\mathbf{x}_1}\!\left[\left\lVert \hat{\mathbf{v}}(\mathbf{x}_t,t,\mathbf{c}_t)-\left(\mathbf{x}_1-\mathbf{x}_0\right)\right\rVert_2^2\right],
\label{eq:method_flow_loss}
\end{equation}
where $\mathbf{c}_t=[\mathbf{p}_t,\mathbf{m}_t]$ serves as the strict state-conditioning vector. During inference, physically viable recovery trajectories are sampled by integrating the learned velocity field $\hat{\mathbf{v}}$ from $t=1$ to $t=0$ utilizing a minimal number of Euler steps.

\textbf{Architecture and Kinematic Continuity.}
The velocity field is parameterized by an AdaLN-modulated Transformer~\cite{peebles2023dit}, where the conditioning vector $\mathbf{c}_t$ and the flow timestep embedding are injected into each block via adaptive shift-scale-gate modulation. To guarantee kinematic continuity between the real-time physical state and the synthesized trajectory, we pin the first residual token via an inpainting constraint during integration:
\begin{equation}
\mathbf{x}_t[0] = (1-t_{\mathrm{next}})\mathbf{r}_0 + t_{\mathrm{next}}\boldsymbol{\epsilon}, \qquad \boldsymbol{\epsilon}\sim\mathcal{N}(0,\mathbf{I}),
\label{eq:method_inpaint}
\end{equation}
where $\mathbf{r}_0$ strictly defines the zero-residual anchor corresponding to the initial state.

\textbf{Receding-Horizon Planning with Directional Warm Start.}
A key design principle is that the middleware always predicts a fixed temporal window of motion, regardless of the distance to the target command. The target is set to the current reference frame $\mathbf{m}_t$, and both the temporal window $\Delta t$ and execution interval $N_{\mathrm{exec}}$ are held constant. Whether the robot is closely tracking the reference or operating far from the reference manifold, the generated keyframes consistently encode the next $\Delta t$ seconds of motion. Long-horizon recovery emerges autoregressively through successive replan cycles, rather than requiring the model to reason about the full trajectory in a single pass.

During inference, we initialize the ODE solver with a directional motion prior rather than pure Gaussian noise. We construct an initial residual trajectory linearly interpolating toward the target:
\begin{equation}
\mathbf{r}^{\mathrm{init}}_k = \frac{k}{K-1}(\mathbf{m}_t - \mathbf{p}_t), \qquad k \in \{0,\ldots,K{-}1\},
\label{eq:method_sdedit_init}
\end{equation}
and begin ODE integration from a partially noised version of this prior at $t_{\mathrm{start}} < 1$:
\begin{equation}
\mathbf{x}_{t_{\mathrm{start}}} = (1 - t_{\mathrm{start}})\, \mathrm{normalize}(\mathbf{r}^{\mathrm{init}}) + t_{\mathrm{start}}\, \boldsymbol{\epsilon}, \qquad \boldsymbol{\epsilon} \sim \mathcal{N}(0, \mathbf{I}),
\label{eq:method_sdedit}
\end{equation}
where $t_{\mathrm{start}}$ controls the noise-to-prior ratio. Inspired by SDEdit~\cite{meng2022sdedit}, this warm start provides the linear prior as an immediate directional estimate, while the learned velocity field refines it into a natural trajectory within a reduced number of ODE steps.
The resulting $K$ keyframes are then densified into the tracker's operating frequency via cubic spline interpolation for joint positions and spherical linear interpolation for root orientations, yielding the complete dense reference signal consumed by the tracking policy.

\subsection{General-Purpose Physics Tracker}
\label{sec:tracker}

\subsubsection{Motion Tracking Formulation}

\textbf{Observation Space.}
The observation space is formally defined as a composite vector $\mathbf{o}_t = \{\mathbf{p}_t,\, \mathbf{m}_t,\, \mathbf{z}_d\}$, comprising the robot's proprioceptive state $\mathbf{p}_t$, the kinematic motion reference trajectory $\mathbf{m}_t$, and the discrete high-level motion embedding $\mathbf{z}_d$. Specifically, the proprioceptive state $\mathbf{p}_t$ encapsulates the immediate physical condition of the robot:
\begin{equation}
\mathbf{p}_t = [\mathbf{g}_t^{\mathrm{proj}},\, \boldsymbol{\omega}_t,\, \mathbf{q}_t - \mathbf{q}_0,\, \dot{\mathbf{q}}_t,\, \mathbf{a}_{t-1}],
\label{eq:method_proprio}
\end{equation}
where $\mathbf{g}_t^{\mathrm{proj}} \in \mathbb{R}^{3}$ denotes the gravity vector projected into the local root frame, $\boldsymbol{\omega}_t \in \mathbb{R}^{3}$ represents the root angular velocity, $\mathbf{q}_t \in \mathbb{R}^{29}$ and $\dot{\mathbf{q}}_t \in \mathbb{R}^{29}$ represent the current joint positions and velocities respectively, $\mathbf{q}_0$ defines the default nominal joint configuration, and $\mathbf{a}_{t-1} \in \mathbb{R}^{29}$ records the previous action to ensure temporal smoothness.

The reference observation $\mathbf{m}_t$ provides per-step target kinematics:
\begin{equation}
\mathbf{m}_t = [\mathbf{v}_t^{\mathrm{ref}},\, \boldsymbol{\omega}_t^{\mathrm{ref}},\, \mathbf{e}_t^{\mathrm{root}},\, \mathbf{q}_t^{\mathrm{ref}}].
\label{eq:method_motion_ref}
\end{equation}
During training, $\mathbf{m}_t$ is extracted directly from the motion dataset. At deployment, it is seamlessly replaced by the densified middleware output $\mathbf{m}'_t$ without any modification to the tracker architecture.
Here, $\mathbf{v}_t^{\mathrm{ref}} \in \mathbb{R}^{3}$ and $\boldsymbol{\omega}_t^{\mathrm{ref}} \in \mathbb{R}^{3}$ denote the reference root linear and angular velocities expressed in the body frame, $\mathbf{q}_t^{\mathrm{ref}} \in \mathbb{R}^{29}$ specifies the target joint positions. The root orientation error $\mathbf{e}_t^{\mathrm{root}} \in \mathbb{R}^{6}$ is strictly parameterized using a 6D continuous rotation feature (Rot6D), computed from the first two columns of the relative rotation matrix $R_{\mathrm{root}}^{\mathrm{des}} R_{\mathrm{root}}^{\top}$. The discrete motion token $\mathbf{z}_d$ captures temporally coherent, high-level motion semantics and is detailed in the subsequent policy architecture description.

\textbf{Action Space.}
The policy $\pi$ outputs target joint positions $\mathbf{a}_t \in \mathbb{R}^{29}$. Each physical joint tracks these respective targets utilizing a low-level Proportional-Derivative (PD) controller operating at high frequency, ensuring stable torque generation.

\textbf{Rewards and Domain Randomization.}
The reward combines positive tracking terms---covering root velocities, body-link orientations, and joint-position matching---with regularization penalties on action jerks, joint-limit violations, and undesired contacts (\cref{tab:teacher_reward_terms}). To ensure sim-to-real transferability, we inject comprehensive domain randomization during training, perturbing both the simulator's physical properties (friction, center-of-mass offsets) and the command-level kinematic targets (\cref{tab:domain_randomization}).

\definecolor{hdr}{RGB}{218,232,252}
\definecolor{altrow}{RGB}{248,250,254}
\definecolor{catbg}{RGB}{240,245,252}
\definecolor{icoTrk}{RGB}{80,140,210}
\definecolor{icoReg}{RGB}{215,135,75}
\definecolor{icoCmd}{RGB}{75,165,145}
\definecolor{icoPhy}{RGB}{145,115,185}
\definecolor{icoDst}{RGB}{205,95,95}
\newcommand{\ico}[1]{\tikz[baseline=-0.55ex]{\fill[#1] (0,0) circle (0.28em);}}
\newcommand{\icosq}[1]{\tikz[baseline=-0.5ex]{\fill[#1,rounded corners=0.6pt] (0,0) rectangle (0.5em,0.5em);}}
\newcommand{\icotri}[1]{\tikz[baseline=-0.4ex]{\fill[#1] (0,0) -- (0.5em,0) -- (0.25em,0.45em) -- cycle;}}

\begin{table*}[!htbp]
\centering
\captionsetup{justification=raggedright,singlelinecheck=false}
\caption{\textbf{Training configuration for the general-purpose physics tracker.} Left: reward function with exponential tracking terms and regularization penalties. Right: domain randomization ranges for command-level, physical, and external disturbance parameters.}
\label{tab:tracker_config}
\vspace{-2pt}
\footnotesize
\setlength{\tabcolsep}{5pt}
\begin{subtable}[t]{0.50\textwidth}
\centering
\caption{Reward function. Tracking: $r_i = w_i\exp\!\bigl(-\lVert e_i\rVert^2\!/\sigma_i^2\bigr)$.}
\label{tab:teacher_reward_terms}
\begin{tabular}{@{}lcc@{}}
\toprule
\rowcolor{hdr}
\textbf{Tracked Quantity} &
$\bm{w}$ &
$\bm{\sigma}$ \\
\midrule
\ico{icoTrk}\; Root Position             & $0.1$ & $0.30$ \\
\rowcolor{altrow}
\ico{icoTrk}\; Root Orientation           & $0.5$ & $0.40$ \\
\ico{icoTrk}\; Root Linear Vel.           & $1.0$ & $0.50$ \\
\rowcolor{altrow}
\ico{icoTrk}\; Root Angular Vel.          & $1.0$ & $1.00$ \\
\ico{icoTrk}\; Rel.\ Body Position        & $1.0$ & $0.30$ \\
\rowcolor{altrow}
\ico{icoTrk}\; Rel.\ Body Orientation     & $1.0$ & $0.40$ \\
\ico{icoTrk}\; Body Linear Vel.           & $1.0$ & $1.00$ \\
\rowcolor{altrow}
\ico{icoTrk}\; Body Angular Vel.          & $1.0$ & $\sqrt{\pi}$ \\
\addlinespace[3pt]
\rowcolor{hdr}
\textbf{Regularization} &
$\bm{w}$ &
\textbf{Penalty} \\
\midrule
\icosq{icoReg}\; Action Rate          & $-0.1$ & $\lVert a_t{-}a_{t\text{-}1}\rVert^2$ \\
\rowcolor{altrow}
\icosq{icoReg}\; Joint Limit          & $-10$  & $\textstyle\sum\max(0,\, q{-}q_{\mathrm{lim}})$ \\
\icosq{icoReg}\; Undesired Contacts   & $-0.1$ & $\textstyle\sum\mathbb{1}(F_c{>}1)$ \\
\bottomrule
\end{tabular}
\end{subtable}%
\hfill
\begin{subtable}[t]{0.47\textwidth}
\centering
\caption{Domain randomization ranges.}
\label{tab:domain_randomization}
\begin{tabular}{@{}ll@{}}
\toprule
\rowcolor{hdr}
\textbf{Parameter} &
\textbf{Range} \\
\midrule
\rowcolor{catbg}
\multicolumn{2}{@{}l}{\icotri{icoCmd}\;\textit{Command-level perturbations}} \\
Target Joint Pos.\ (rad)      & $\pm\, 0.01$ \\
\rowcolor{altrow}
Target Ang.\ Vel.\ (rad/s)    & $\pm\, 0.2$ \\
Target Root Rot.\ (rad)       & $\pm\, 0.05$ \\
\rowcolor{altrow}
Default Joint Pos.\ (rad)     & $\pm\, 0.01$ \\
\addlinespace[2pt]
\rowcolor{catbg}
\multicolumn{2}{@{}l}{\ico{icoPhy}\;\textit{Physical properties}} \\
CoM Offset (m)                & $x{:}\,\pm 0.5;\; y,z{:}\,\pm 0.1$ \\
\rowcolor{altrow}
Static Friction               & $[0.3,\; 1.6]$ \\
Dynamic Friction              & $[0.3,\; 1.2]$ \\
\addlinespace[2pt]
\rowcolor{catbg}
\multicolumn{2}{@{}l}{\icosq{icoDst}\;\textit{External disturbances}} \\
Push Frequency (s)            & $1.0$\,--\,$3.0$ \\
\rowcolor{altrow}
Push Lin.\ Vel.\ (m/s)        & $xy{:}\,\pm 0.5;\; z{:}\,\pm 0.2$ \\
Push Ang.\ Vel.\ (rad/s)      & $RP{:}\,\pm 0.52;\; Y{:}\,\pm 0.78$ \\
\bottomrule
\end{tabular}
\end{subtable}
\end{table*}

\subsubsection{General Motion Tracking Policy}

Our tracking policy is built on a shared motion-latent representation with an encoder--quantizer--decoder structure. A motion encoder maps the kinematic observations $\mathbf{m}_t$ to a continuous latent $\mathbf{z}_c$, which is then discretized into tokenized codes $\mathbf{z}_d$. Two parallel heads consume $\mathbf{z}_d$: a reconstruction decoder for representation learning, and an action decoder that fuses $\mathbf{z}_d$ with the proprioceptive state $\mathbf{p}_t$ to produce control actions.

\textbf{Improved Discrete Quantization.}
We adopt an improved Finite Scalar Quantization (iFSQ)~\cite{lin2026ifsqimprovingfsqimage} to distill high-frequency kinematic signals into compact semantic tokens. Given continuous latent features $\mathbf{z}_c \in \mathbb{R}^{N \times d}$, with $N$ the batch size and $d$ the embedding dimension, each channel is bounded to $[-1,1]$ and quantized into $L=2^K+1$ uniformly spaced levels, where the extra center level guarantees an exact zero-state. Element-wise quantization maps each dimension to an integer index $z_d \in \{0,\ldots,L-1\}$ via:
\begin{equation}
z_d = \mathrm{round}\!\left(\frac{L-1}{2}\left(f(z_c)+1\right)\right).
\label{eq:method_fsq_quant}
\end{equation}
Rather than the standard $\tanh$ bounding, we employ a sigmoid-based mapping that improves bin utilization while preserving uniform quantization intervals:
\begin{equation}
f(x) = 2.0\,\sigma(1.6x) - 1.
\label{eq:method_ifsq_bound}
\end{equation}
We apply the straight-through estimator (stop-gradient) for the rounding operation, yielding the discrete tokens $\mathbf{z}_d$ injected into the policy observation.

\textbf{Encoder-Decoder Architecture.}
The motion encoder ingests a 10-frame future window $\mathbf{M}_{t:t+9}$ and produces a continuous embedding $\mathbf{z}_c \in \mathbb{R}^{d}$. After iFSQ discretization, the reconstruction decoder maps $\mathbf{z}_d$ back to the full 10-frame motion sequence $\hat{\mathbf{M}}_{t:t+9}$, optimized via:
\begin{equation}
\mathcal{L}_{\mathrm{rec}} = \frac{1}{10}\sum_{k=0}^{9}\left\lVert \hat{\mathbf{m}}_{t+k} - \mathbf{m}_{t+k} \right\rVert_2^2.
\label{eq:method_rec_mse}
\end{equation}
The action decoder concatenates $\mathbf{z}_d$ with a 10-step proprioceptive history $\mathbf{P}_{t-9:t}$ to produce the control action $\mathbf{a}_t$.

\textbf{Adaptive Motion Sampling.}
Training a unified policy over a large, heterogeneous motion corpus introduces severe optimization imbalance: uniform sampling overfits to trivial locomotion while underperforming on dynamic, agile skills. Inspired by the bin-level adaptive curriculum in ZEST~\cite{sleiman2026zest}, we design a continuous temporal-bin variant with smoothed difficulty propagation.

We partition the concatenated motion corpus into $B$ uniformly spaced temporal bins. At each episode termination, the per-step tracking score $s_t \in [0,1]$ is averaged over the episode to yield a difficulty estimate $d = 1 - \bar{s}$, which is accumulated into the corresponding bin via exponential moving average:
\begin{equation}
F_b \leftarrow \alpha \, d_b + (1 - \alpha) \, F_b,
\label{eq:method_bin_ema}
\end{equation}
where $F_b$ denotes the difficulty of bin $b$ and $\alpha$ controls the update rate. To prevent isolated hard bins from dominating the sampling distribution and to propagate difficulty to temporally adjacent regions, we apply a 1D kernel smoothing operation followed by an outlier cap:
\begin{equation}
\tilde{F}_b = \left(\mathcal{K} * \hat{F}\right)_b, \quad \hat{F}_b = \min\!\left(F_b, \, c \cdot \bar{F}\right),
\label{eq:method_kernel_smooth}
\end{equation}
where $\mathcal{K}$ is an exponentially decaying kernel and $c$ bounds the maximum bin weight relative to the global mean $\bar{F}$. The final sampling distribution mixes the smoothed difficulty with a uniform baseline to ensure exploration:
\begin{equation}
P_b = \eta \, \tilde{F}_b + \frac{1 - \eta}{B},
\label{eq:method_adaptive_sample}
\end{equation}
where $\eta$ controls the balance between difficulty-driven and uniform sampling. This mechanism continuously steers the training distribution toward challenging temporal regions of the motion manifold, while the kernel smoothing ensures that difficulty information propagates to neighboring segments, preventing abrupt sampling discontinuities.

\subsection{Training Paradigm for Unified Tracking and Generation}

\textbf{Dataset Construction.}
Training tuples for the state-conditioned middleware are generated from a diverse motion corpus using a receding-horizon sampling strategy. For each motion sequence, segment starting points are selected at regular intervals, and the temporal segment length $\ell$ is drawn from a log-uniform distribution:
\begin{equation}
\ell \sim \exp\!\left(\mathcal{U}(\log \ell_{\min},\, \log \ell_{\max})\right),
\label{eq:method_seg}
\end{equation}
where $\ell_{\min} = H$ is set equal to the planning horizon. From each segment, $K$ uniformly spaced keyframes are extracted covering only the first $H$ frames (corresponding to $\Delta t = H/\text{fps}$ seconds), regardless of the total segment length. The conditioning vector $\mathbf{c}_t = [\tilde{\mathbf{p}}_t,\, \mathbf{m}_t]$ pairs the start state with the reference command at the segment endpoint. The residual supervision is computed against the static baseline $\boldsymbol{\beta}_{t,k} = \tilde{\mathbf{p}}_t$ (\cref{eq:method_residual}), so each training target encodes the motion increment over the next $\Delta t$ seconds from the current state. This design ensures that (i) the model never needs to predict trajectories exceeding a fixed temporal horizon, bounding the residual magnitude regardless of the state-command gap; (ii) training naturally covers every phase of long recovery sequences, as successive starting points within the same motion yield overlapping local windows; and (iii) eliminating near-zero-length segments ($\ell < H$) removes the trivial zero-residual bias that otherwise dominates under mean-squared-error training.

\textbf{Noisy-State Augmentation.}
Deployment introduces a systematic discrepancy between noisy physical state estimation and the clean reference commands available during training. To close this gap, we apply asymmetric start-state perturbations: only the initial proprioceptive state is corrupted with channel-wise Gaussian noise,
\begin{equation}
\tilde{\mathbf{p}}_t = \mathbf{p}_t + \boldsymbol{\epsilon},
\label{eq:method_noise}
\end{equation}
while the target reference command remains clean. The noise magnitude is scaled per channel to reflect the varying sensitivity of different state dimensions. This asymmetric augmentation mirrors the deployment scenario where the robot's proprioceptive state reflects both sensor noise and accumulated tracking errors, while the reference command is always clean, improving robustness to real-world state-command discrepancies.

\textbf{Kinematics-Aware Loss Weighting.}
Beyond the primary velocity matching objective (\cref{eq:method_flow_loss}), we introduce a kinematics-aware weighting scheme motivated by the observation that identical joint-space errors can induce vastly different body-space deviations depending on the current kinematic configuration. Concretely, we weight each state dimension $q_d$ by a pose-dependent Jacobian magnitude:
\begin{equation}
w_d(\mathbf{q}) = \sum_b \left\lVert \frac{\partial \mathbf{p}_b}{\partial q_d} \right\rVert_2^2
\;\approx\;
\sum_b \left\lVert \frac{\mathbf{p}_b(\mathbf{q}+\delta e_d) - \mathbf{p}_b(\mathbf{q}-\delta e_d)}{2\delta} \right\rVert_2^2,
\label{eq:method_jac}
\end{equation}
where $\mathbf{p}_b$ is the Cartesian position of tracked body link $b$. This approximates the diagonal of $\mathbf{J}^{\top}\mathbf{J}$ evaluated independently at each training pose and keyframe, producing a weight tensor of shape $(N, K, D)$. This captures pose-dependent lever arm geometry: for example, a shoulder joint in a T-pose configuration commands a longer moment arm than when the arm hangs at rest, and receives a correspondingly larger gradient signal. The weights are normalized to unit mean per dimension and clamped to a minimum value to prevent zero-gradient dimensions. In practice, all weights are precomputed via differentiable forward kinematics during dataset construction and cached alongside training tuples, eliminating all online overhead.

\textbf{State Representation and Model Variants.}
We evaluate two state parameterizations: a 38D configuration comprising joint positions (29D), root position (3D), and root orientation in 6D continuous rotation representation, and a 35D variant omitting the global root position. The 38D formulation enables the middleware to synthesize root translational commands, allowing the robot to autonomously correct global positional drift relative to the reference trajectory. The 35D variant delegates root translation entirely to the reference motion, decoupling the middleware from global localization. While this sacrifices autonomous position correction, it eliminates dependence on external positioning systems, making it directly deployable with onboard proprioception and IMU alone. Both configurations retain equivalent fall recovery and general motion tracking performance. All quantitative results reported in this work use the 38D configuration unless otherwise noted.

%% file: sections/4_experiments_and_results.tex
\section{Experiments}

\definecolor{bestblue}{RGB}{218,232,252}
\newcommand{\pctg}[1]{{\tiny\textcolor{teal!80!black}{(#1\%)}}}
\newcommand{\pctb}[1]{{\tiny\textcolor{red!70!black}{(#1\%)}}}

\subsection{Implementation Details}

\textbf{Simulation Environment.}
All experiments are conducted on the Unitree G1 humanoid platform, a full-size bipedal robot standing approximately 1.32\,m tall with a total mass of roughly 35\,kg. The robot features 29 actuated degrees of freedom spanning the torso, two 7-DoF arms, and two 6-DoF legs, all driven by proprietary electric actuators. Training is carried out in IsaacLab~\cite{mittal2025isaac}, a GPU-accelerated simulator built on NVIDIA Isaac Sim, where we instantiate 16{,}384 parallel environments on a single NVIDIA A100 (80\,GB) GPU. The physics simulation runs at a 5\,ms timestep (200\,Hz) on flat ground with randomized friction coefficients (\cref{tab:domain_randomization}), while the control policy queries observations and emits actions at 50\,Hz (every 4 simulation substeps). Each action is converted to joint torques by a per-joint PD controller executing at the full 200\,Hz rate. For evaluation, all policies are tested in the MuJoCo physics engine on a held-out set comprising 101 unseen motion sequences that span locomotion, dance, martial arts, daily activities, fall-and-recovery, acrobatic jumps, and discretized motion clips in which continuous reference trajectories are replaced with piecewise-constant signals consisting of static poses separated by abrupt transitions, thereby removing all smooth interpolation and testing the policy's ability to track discontinuous commands. Each sequence is rolled out for its full duration (up to 20\,s).

\textbf{Motion Dataset.}
The training corpus is curated from diverse, complementary sources, comprising selected clips from LAFAN1~\cite{harvey2020robust}, 100STYLE~\cite{mason2022realtimestylemodellinghuman}, SnapMoGen~\cite{guo2025snapmogenhumanmotiongeneration}, AMASS~\cite{mahmood2019amass}, and proprietary in-house motion capture recordings. The assembled dataset spans locomotion, martial arts, dance, daily activities, fall-and-recovery sequences, and jumping motions, thereby providing broad stylistic and temporal coverage for evaluating tracking fidelity, robustness, and execution stability under heterogeneous motion commands.

\textbf{Tracker Training.}
The general-purpose physics tracker is trained via Proximal Policy Optimization (PPO) within an asymmetric actor--critic framework, following established practices in prior whole-body tracking systems~\cite{li2025bfmzero,liao2025beyondmimic,luo2025sonic}. The iFSQ encoder, reconstruction decoder, and action decoder are trained jointly end-to-end with the PPO objective, where the reconstruction loss (\cref{eq:method_rec_mse}) and the RL policy gradient share the same encoder--quantizer pathway. The adaptive motion sampling curriculum (\cref{eq:method_adaptive_sample}) is activated after an initial warm-up phase to allow early-stage uniform coverage. The complete reward function and domain randomization ranges are specified in \cref{tab:tracker_config}; all remaining hyperparameters are listed in \cref{tab:tracker_hyperparams}.

\textbf{Trajectory Generator Training.}
The state-conditioned flow matching trajectory generator is trained offline on paired trajectory tuples extracted from the motion corpus following the dataset construction procedure described in \cref{eq:method_seg}. The velocity field $\hat{\mathbf{v}}$ is parameterized by an AdaLN-modulated Transformer~\cite{peebles2023dit} that predicts $K{=}8$ uniformly spaced keyframes over a fixed planning horizon of $\Delta t {=} 0.2$\,s. We train two state representation variants: a \textbf{38D} configuration encoding 29 joint positions, 3D root position, and 6D root orientation, and a \textbf{35D} variant that omits root position entirely. The 38D model retains full awareness of global positioning, enabling recovery toward the reference command manifold in both pose and location. The 35D model relies solely on joint encoders and an IMU for root orientation, making it directly deployable on hardware without external localization; global position tracking is delegated to the reference motion source. We optimize the velocity matching objective (\cref{eq:method_flow_loss}) jointly with the kinematics-aware loss weighting (\cref{eq:method_jac}). Noisy-state augmentation (\cref{eq:method_noise}) is applied throughout training with channel-wise Gaussian noise with reduced magnitudes for root pose and orientation channels. During deployment, trajectory samples are generated via 5 Euler integration steps from $t{=}0.9$ to $t{=}0$, initialized with the directional warm start (\cref{eq:method_sdedit}) at $t_{\mathrm{start}}{=}0.9$. The generator replans every $N_{\mathrm{exec}}{=}2$ control steps (0.04\,s), yielding a closed-loop replanning frequency of 25\,Hz. Full hyperparameters for both the tracker and the trajectory generator are listed in \cref{tab:tracker_hyperparams,tab:generator_hyperparams}.

\begin{table*}[!htbp]
\centering
\captionsetup{justification=raggedright,singlelinecheck=false}
\caption{\textbf{Training hyperparameters for the physics tracker and the trajectory generator.} Left: PPO-based tracker training configuration. Right: flow matching trajectory generator architecture, optimization, and inference settings.}
\vspace{-2pt}
\footnotesize
\begin{minipage}[t]{0.46\linewidth}
\centering
\label{tab:tracker_hyperparams}
\begin{tabular}{@{}ll@{}}
\toprule
\rowcolor{bestblue}
\textbf{Tracker Hyperparameter} & \textbf{Value} \\
\midrule
Parallel environments & 16{,}384 \\
\rowcolor{altrow}
Rollout horizon & 24 steps \\
Discount factor $\gamma$ & 0.99 \\
\rowcolor{altrow}
GAE $\lambda$ & 0.95 \\
PPO epochs / mini-batches & 5 / 4 \\
\rowcolor{altrow}
Clipping ratio $\epsilon$ & 0.2 \\
Actor learning rate & $2\times10^{-3}$ \\
\rowcolor{altrow}
Critic learning rate & $1\times10^{-3}$ \\
KL target & 0.01 \\
\rowcolor{altrow}
Entropy coefficient & 0.005 \\
Gradient clip norm & 1.0 \\
Total training iterations & ${\sim}$100{,}000 \\
\bottomrule
\end{tabular}
\end{minipage}%
\hfill%
\begin{minipage}[t]{0.52\linewidth}
\centering
\label{tab:generator_hyperparams}
\begin{tabular}{@{}ll@{}}
\toprule
\rowcolor{bestblue}
\textbf{Generator Hyperparameter} & \textbf{Value} \\
\midrule
Attention blocks / heads / dim & 6 / 4 / 512 \\
\rowcolor{altrow}
Conditioning injection & AdaLN ($\mathbf{c}_t{=}[\mathbf{p}_t, \mathbf{m}_t]$ + timestep) \\
Keyframes $K$ / horizon $\Delta t$ & 8 / 0.2\,s \\
\rowcolor{altrow}
State dimension $D$ & 38 / 35 \\
Optimizer & AdamW ($\beta_1{=}0.9$, $\beta_2{=}0.999$) \\
\rowcolor{altrow}
Learning rate & $1\times10^{-4}$ (cosine decay) \\
Weight decay & $10^{-4}$ \\
\rowcolor{altrow}
Batch size & 256 \\
Training epochs & 4{,}000 \\
\rowcolor{altrow}
Parameters & 22.9\,M \\
Inference ODE steps & 5 (Euler, $t{:}\,0.9{\to}0$) \\
\rowcolor{altrow}
Warm-start $t_{\mathrm{start}}$ & 0.9 \\
Replan interval $N_{\mathrm{exec}}$ & 2 steps (0.04\,s, 25\,Hz) \\
\bottomrule
\end{tabular}
\end{minipage}
\end{table*}

\FloatBarrier
\subsection{Comparisons}

We evaluate eight model configurations that systematically vary four design axes---policy architecture, motion tokenizer, observation design, and generative trajectory planning---to isolate the contribution of each component. All variants share the same simulator, robot morphology, reward function, and training budget, and are evaluated on the identical held-out motion set.

\textbf{Compared Methods.}
All policies receive a 10-frame proprioceptive history and a 10-frame future motion reference window (frames $t$ through $t{+}9$) unless otherwise noted. We organize the evaluated methods into two groups; detailed network architecture specifications are provided in \cref{tab:arch_details}.

\textit{Architecture and external baselines.}
\textbf{MLP} employs a standard MLP actor--critic with BeyondMimic-style (BM) observations~\cite{liao2025beyondmimic}.
\textbf{Transformer} follows the TokenHSI architecture~\cite{pan2025tokenhsi}, encoding proprioceptive and motion inputs into per-modality tokens that are processed by a multi-head Transformer with a learnable aggregation token before an MLP action head.
\textbf{VQ-VAE} replaces the iFSQ quantizer with a VQ-VAE tokenizer~\cite{sun2024prompt} while retaining the same encoder--decoder policy structure.
\textbf{SONIC}~\cite{luo2025sonic} is reproduced from its official open-source release with the original observation design.

\textit{Heracles variants.}
$\textbf{iFSQ}_{\textbf{BM}}$ pairs the iFSQ tokenizer and encoder--decoder policy with BeyondMimic observations augmented by height features (BM+H), isolating the tokenizer contribution under a standard observation design.
$\textbf{iFSQ}_{\textbf{+H}}$ combines the iFSQ-based policy with the proposed observation design (\cref{sec:tracker}) including explicit height features.
\textbf{iFSQ} uses the proposed observation design without height, representing the best standalone tracker configuration.
\textbf{Heracles} augments the iFSQ tracker with the state-conditioned trajectory generator (\cref{eq:method_middleware}), constituting the full proposed system.

\textbf{Evaluation Protocol.}
All methods are evaluated on the same held-out set of 101 motion sequences unseen during training, spanning locomotion, dance, martial arts, daily activities, and fall-and-recovery. Each sequence is rolled out for its full duration (up to 20\,s). We report five metrics: (i)~\textit{Completion Rate (CR)}---the fraction of reference frames for which the policy maintains a root height error below 0.3\,m and a root orientation error below 1.2\,rad; (ii)~\textit{Joint Position Error}---the $L_2$ norm of joint-position deviations; (iii)~\textit{Root Height Error}---absolute height deviation in the world frame; (iv)~\textit{Root Orientation Error}---orientation error excluding yaw; and (v)~\textit{Root Linear Velocity Error}---velocity error in the body frame.

\begin{table*}[!htbp]
\centering
\captionsetup{justification=raggedright,singlelinecheck=false}
\caption{\textbf{Network architecture specifications for all evaluated methods.} $[\cdot]$ denotes MLP hidden-layer widths. iFSQ$_{*}$ covers all iFSQ variants (iFSQ$_{\text{BM}}$, iFSQ$_{\text{+H}}$, iFSQ), which share the same network but differ in observation design.}
\label{tab:arch_details}
\vspace{-2pt}
\footnotesize
\begin{tabular}{@{}lll@{}}
\toprule
\rowcolor{bestblue}
\textbf{Method} & \textbf{Component} & \textbf{Configuration} \\
\midrule
MLP              & Actor       & $[4096, 2048, 1024, 512, 256]$ MLP \\
                 & Critic      & $[4096, 2048, 1024, 512, 256]$ MLP \\
\addlinespace[3pt]
\rowcolor{altrow}
Transformer      & Tokenizer   & $[512, 512]$ MLP $\times\,2$ modalities $\to$ 3 tokens (512-dim) \\
\rowcolor{altrow}
                 & Backbone    & 3-layer, 4-head, 512-dim Transformer \\
\rowcolor{altrow}
                 & Action head & $[2048, 1024, 256]$ MLP \\
\rowcolor{altrow}
                 & Critic      & $[3072, 1536, 768, 512]$ MLP \\
\addlinespace[3pt]
VQ-VAE           & Quantizer   & Codebook $|\mathcal{C}|{=}10{,}240$, dim${=}512$ \\
                 & Policy      & Enc-Dec (identical to iFSQ) \\
\addlinespace[3pt]
\rowcolor{altrow}
SONIC            & --          & Official release~\cite{luo2025sonic}; stride-5 reference sampling \\
\addlinespace[3pt]
iFSQ$_{*}$       & Policy      & iFSQ Enc-Dec (\cref{sec:tracker}) \\
\addlinespace[3pt]
\rowcolor{altrow}
Heracles         & Tracker     & iFSQ Enc-Dec (\cref{sec:tracker}) \\
\rowcolor{altrow}
                 & Traj.\ gen. & 6-layer, 4-head, 512-dim AdaLN Transformer~\cite{peebles2023dit} \\
\bottomrule
\end{tabular}
\end{table*}

\begin{table*}[!htbp]
\captionsetup{justification=raggedright,singlelinecheck=false}
\caption{\textbf{Component configuration and quantitative comparison of all evaluated methods.} \textbf{(a)}~Obs.\ column: BM\,=\,BeyondMimic-style~\cite{liao2025beyondmimic}; $\dagger$\,=\,proposed (\cref{sec:tracker}); +H\,=\,with height features. \cmark/\xmark{} indicates presence or absence of the trajectory generator. 
\textbf{(b)}~Completion rate measures the fraction of reference frames for which the root height error remains below 0.3\,m and the root orientation error stays below 1.2\,rad; tracking errors are reported for joint positions, root height, root orientation (excl.\ yaw), and root linear velocity.
\colorbox{bestblue}{Blue} cells mark the best result per metric. Colored percentages show relative change from MLP: \textcolor{teal!80!black}{teal}\,=\,better, \textcolor{red!70!black}{red}\,=\,worse.}
\label{tab:comparison}
\vspace{-2pt}
\footnotesize
\centering
\textbf{(a)} Component configuration \label{tab:method_config}
\vspace{2pt}

\begin{tabular}{@{}llccc@{}}
\toprule
\textbf{Method} & \textbf{Obs.} & \textbf{Tokenizer} & \textbf{Policy} & \textbf{Traj.\ Gen.} \\
\midrule
MLP              & BM             & --   & MLP     & \xmark \\
Transformer      & BM             & --   & Trans.  & \xmark \\
VQ-VAE           & BM             & VQ   & Enc-Dec & \xmark \\
SONIC            & SONIC          & --   & SONIC   & \xmark \\
\addlinespace[3pt]
iFSQ$_{\text{BM}}$  & BM+H       & iFSQ & Enc-Dec & \xmark \\
iFSQ$_{\text{+H}}$  & $\dagger$+H & iFSQ & Enc-Dec & \xmark \\
iFSQ             & $\dagger$      & iFSQ & Enc-Dec & \xmark \\
Heracles         & $\dagger$      & iFSQ & Enc-Dec & \cmark \\
\bottomrule
\end{tabular}

\vspace{8pt}
\textbf{(b)} Tracking performance \label{tab:tracking_perf}
\vspace{2pt}

\begin{tabular}{@{}lccccc@{}}
\toprule
\textbf{Method}
  & \makecell{\textbf{CR}\\(\%) $\uparrow$}
  & \makecell{\textbf{Joint Err}\\(rad) $\downarrow$}
  & \makecell{\textbf{Height Err}\\(m) $\downarrow$}
  & \makecell{\textbf{Ori Err}\\(rad) $\downarrow$}
  & \makecell{\textbf{LinVel Err}\\(m/s) $\downarrow$} \\
\midrule
MLP              & $84.8$ & \cellcolor{bestblue}$\mathbf{1.1572}$ & $0.1194$ & $0.3590$ & $0.2230$ \\
\rowcolor{altrow}
Transformer      & $80.6$\,{\color{red}\scriptsize($-5.0\%$)} & $1.5436$\,{\color{red}\scriptsize($-33.4\%$)} & $0.1426$\,{\color{red}\scriptsize($-19.4\%$)} & $0.3410$\,{\color{teal}\scriptsize($+5.0\%$)} & $0.2046$\,{\color{teal}\scriptsize($+8.3\%$)} \\
VQ-VAE           & $86.0$\,{\color{teal}\scriptsize($+1.4\%$)} & $2.3013$\,{\color{red}\scriptsize($-98.9\%$)} & $0.1077$\,{\color{teal}\scriptsize($+9.8\%$)} & $0.3675$\,{\color{red}\scriptsize($-2.4\%$)} & $0.2376$\,{\color{red}\scriptsize($-6.5\%$)} \\
\rowcolor{altrow}
SONIC            & $79.3$\,{\color{red}\scriptsize($-6.5\%$)} & $1.9828$\,{\color{red}\scriptsize($-71.3\%$)} & $0.1402$\,{\color{red}\scriptsize($-17.4\%$)} & $0.3771$\,{\color{red}\scriptsize($-5.0\%$)} & $0.2334$\,{\color{red}\scriptsize($-4.7\%$)} \\
\addlinespace[3pt]
iFSQ$_{\text{BM}}$  & $85.1$\,{\color{teal}\scriptsize($+0.4\%$)} & $1.4760$\,{\color{red}\scriptsize($-27.5\%$)} & $0.1096$\,{\color{teal}\scriptsize($+8.2\%$)} & $0.3474$\,{\color{teal}\scriptsize($+3.2\%$)} & $0.2362$\,{\color{red}\scriptsize($-5.9\%$)} \\
\rowcolor{altrow}
iFSQ$_{\text{+H}}$  & $87.3$\,{\color{teal}\scriptsize($+2.9\%$)} & $1.2924$\,{\color{red}\scriptsize($-11.7\%$)} & \cellcolor{bestblue}$\mathbf{0.0271}$\,{\color{teal}\scriptsize($+77.3\%$)} & \cellcolor{bestblue}$\mathbf{0.1539}$\,{\color{teal}\scriptsize($+57.1\%$)} & $0.1709$\,{\color{teal}\scriptsize($+23.4\%$)} \\
iFSQ             & $87.2$\,{\color{teal}\scriptsize($+2.8\%$)} & $1.1863$\,{\color{red}\scriptsize($-2.5\%$)} & $0.0955$\,{\color{teal}\scriptsize($+20.0\%$)} & $0.3614$\,{\color{red}\scriptsize($-0.7\%$)} & \cellcolor{bestblue}$\mathbf{0.1561}$\,{\color{teal}\scriptsize($+30.0\%$)} \\
\rowcolor{altrow}
Heracles         & \cellcolor{bestblue}$\mathbf{90.6}$\,{\color{teal}\scriptsize($+6.8\%$)} & $1.3272$\,{\color{red}\scriptsize($-14.7\%$)} & $0.0764$\,{\color{teal}\scriptsize($+36.0\%$)} & $0.2728$\,{\color{teal}\scriptsize($+24.0\%$)} & $0.2325$\,{\color{red}\scriptsize($-4.3\%$)} \\
\bottomrule
\end{tabular}
\end{table*}

\textbf{Results.}
\cref{tab:method_config}(a) summarizes the component configuration of each variant, and quantitative tracking performance is reported in \cref{tab:tracking_perf}(b). We highlight four principal findings.

\textit{Robustness.}
The three variants equipped with the proposed observation design and iFSQ tokenizer (iFSQ$_{\text{+H}}$, iFSQ, Heracles) consistently outperform all baselines in completion rate, achieving 87.3\%, 87.2\%, and 90.6\% respectively. Heracles attains the highest completion rate, exceeding the best external baseline VQ-VAE (86.0\%) by 4.6 percentage points and MLP (84.8\%) by 5.8 points. Among external baselines, completion rates range from 79.3\% (SONIC) to 86.0\% (VQ-VAE). Switching from BM to the proposed observation design while keeping the iFSQ tokenizer fixed (iFSQ$_{\text{BM}}$ $\to$ iFSQ) raises completion from 85.1\% to 87.2\%, confirming the role of observation design in robust tracking.

\textit{Tokenizer effectiveness.}
Comparing VQ-VAE and iFSQ$_{\text{BM}}$---which share the same encoder--decoder policy and BM-style observation design but differ in the quantizer---reveals that iFSQ reduces the joint-position error from 2.3013 to 1.4760\,rad ($-$35.8\%) while maintaining a comparable completion rate (85.1\% vs.\ 86.0\%). The dramatic reduction in tracking precision highlights the superior codebook utilization of finite scalar quantization over conventional VQ-VAE in this high-frequency control domain.
\begin{figure}[!htbp]
\centering
\includegraphics[width=0.9\columnwidth]{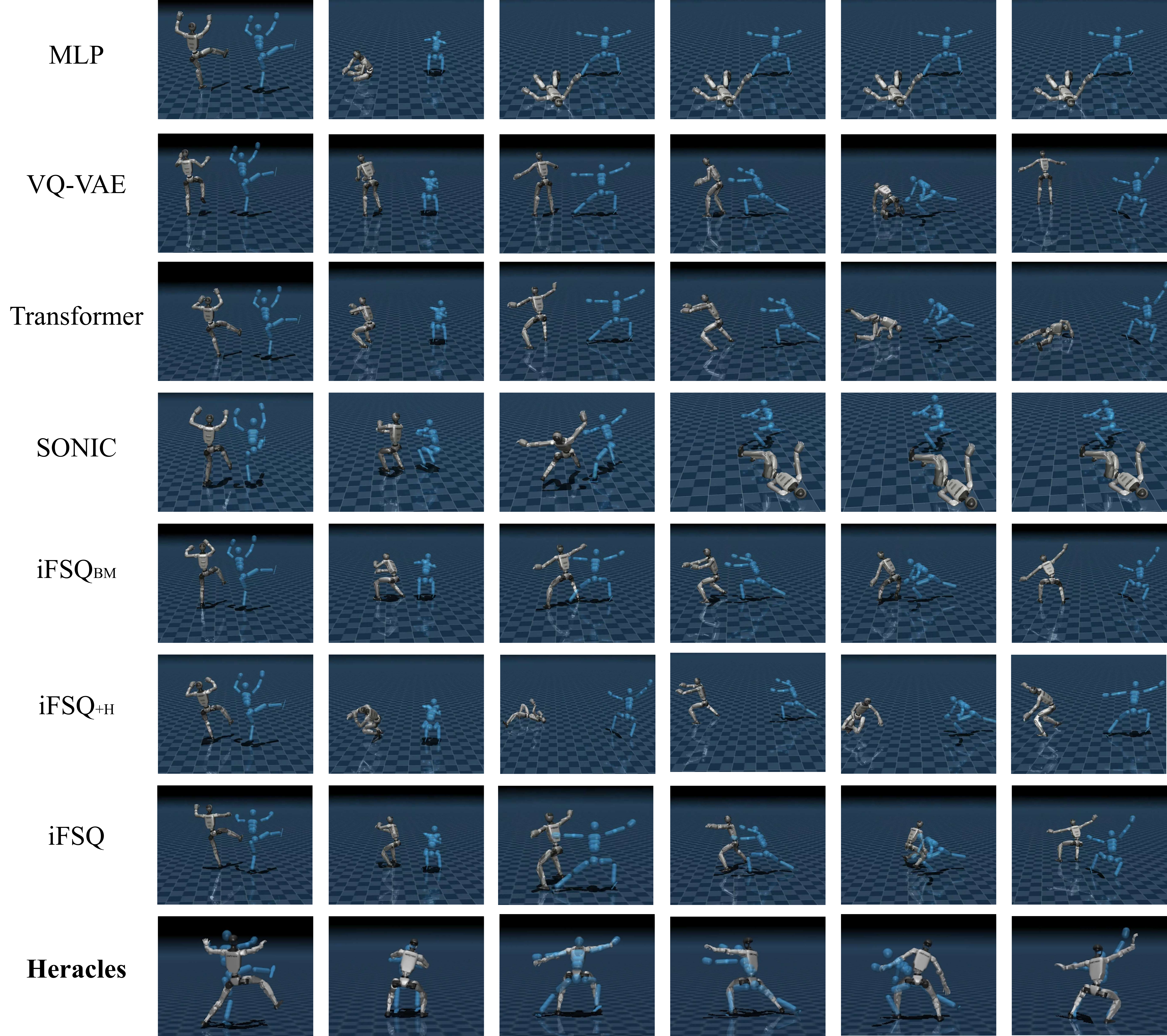}
\captionsetup{justification=raggedright,singlelinecheck=false}
\caption{\textbf{Qualitative sim-to-sim comparison on an out-of-distribution martial arts sequence.} Each row shows a different method tracking the same reference motion on a Unitree G1 humanoid alongside a reference ghost in MuJoCo. MLP, Transformer, and SONIC collapse early; VQ-VAE barely tracks the motion. Among our ablations, iFSQ\textsubscript{BM} and iFSQ survive without falling, while iFSQ\textsubscript{+H} falls but later recovers. \textbf{Heracles} (Ours) tracks the full sequence most accurately, demonstrating the strongest robustness to OOD motions.}
\label{fig:arts}
\end{figure}
\textit{Observation design and height features.}
Among the iFSQ variants, iFSQ$_{\text{+H}}$ achieves the lowest height error (0.0271\,m) and orientation error (0.1539\,rad) across all methods, while iFSQ attains the best linear-velocity tracking (0.1561\,m/s) and a competitive joint-position error (1.1863\,rad). Removing explicit height features (iFSQ$_{\text{+H}}$ $\to$ iFSQ) yields lower joint error (1.1863 vs.\ 1.2924\,rad) at the expense of higher height error (0.0955 vs.\ 0.0271\,m) and markedly degraded orientation control (0.3614 vs.\ 0.1539\,rad), confirming that the height channel is critical for vertical and orientation precision.
\textit{Trajectory generator.}
Heracles achieves the highest completion rate (90.6\%) among all methods---a 6.8\% relative improvement over MLP and a 3.9\% improvement over the standalone iFSQ tracker---while maintaining competitive tracking quality. Compared to iFSQ, the trajectory generator reduces root orientation error from 0.3614 to 0.2728\,rad (24.5\% relative reduction) and height error from 0.0955 to 0.0764\,m (20.0\% reduction), at a modest cost in joint-position error (1.3272 vs.\ 1.1863\,rad). This indicates that the state-conditioned generative planner synthesizes spatially-aware recovery trajectories that refine both vertical and heading control, a capability absent in the reactive tracker alone.

We present qualitative sim-to-sim evaluation results on an out-of-distribution martial arts sequence in MuJoCo, as shown in \cref{fig:arts}. Among the baseline methods, MLP, Transformer, and SONIC fail to maintain balance and collapse early in the sequence, while VQ-VAE barely tracks the motion throughout. For our ablation variants, iFSQ$_{\text{BM}}$ and iFSQ successfully survive the entire sequence without falling; however, iFSQ$_{+\text{H}}$ experiences a fall at an intermediate stage but manages to recover afterwards. In contrast, \textbf{Heracles} not only survives the full sequence but also accurately tracks the root position and body pose across all frames, demonstrating the strongest robustness to out-of-distribution motions among all evaluated approaches.

\begin{figure}[!htbp]
\centering
\includegraphics[width=\columnwidth]{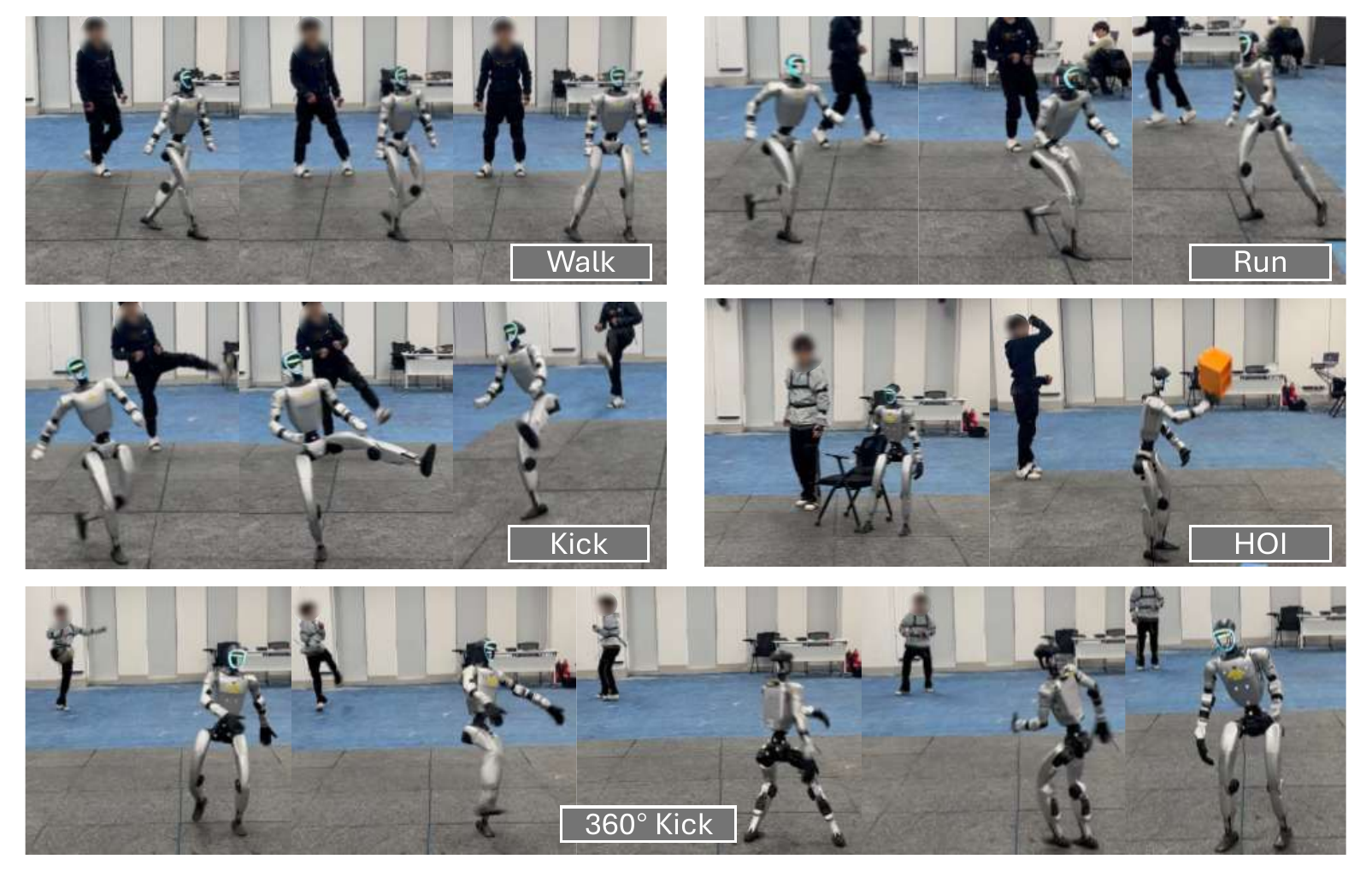}
\captionsetup{justification=raggedright,singlelinecheck=false}
\caption{\textbf{Real-world motion tracking across diverse and dynamic behaviors.} Real-world experiments demonstrate that our model generalizes to a broad spectrum of motions, from everyday locomotion (walk, run) to highly dynamic skills (kick, 360° kick) and human-object interaction.}
\label{fig:realtrack}
\end{figure}

We further validate our method through real-world deployment on a Unitree G1 humanoid robot, as illustrated in \cref{fig:realtrack}. The experiments span a broad spectrum of behaviors, ranging from everyday locomotion such as walking and running, to highly dynamic skills including kicking and full 360° kicks, as well as human-object interaction (HOI) scenarios.

\FloatBarrier
\subsection{Fall-and-Recovery Evaluation}
\label{sec:getup}

\begin{figure}[!htbp]
\centering
\includegraphics[width=0.9\columnwidth]{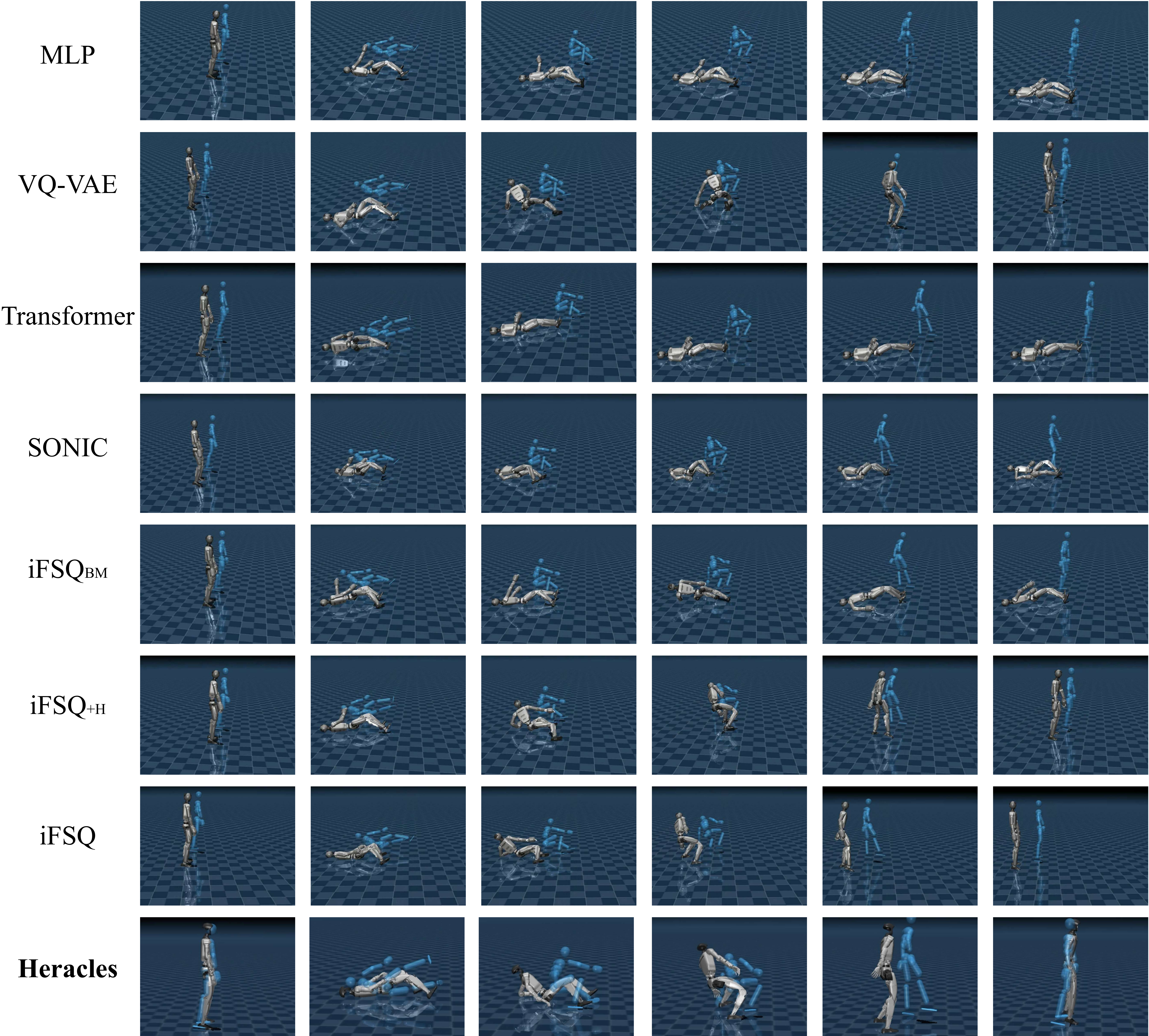}
\captionsetup{justification=raggedright,singlelinecheck=false}
\caption{\textbf{Qualitative sim-to-sim comparison on an OOD lie-to-stand sequence.} Same setup as \cref{fig:arts}. MLP, Transformer, SONIC, and iFSQ\textsubscript{BM} fail to stand up; VQ-VAE, iFSQ\textsubscript{+H}, and iFSQ partially track the motion. \textbf{Heracles} (Ours) completes the full lie-to-stand transition and most accurately tracks the root position.}
\label{fig:lie}
\end{figure}

To specifically assess the fall-and-recovery capabilities that distinguish Heracles from pure tracking approaches, we conduct a dedicated evaluation on a curated subset of fall-and-recovery motion sequences extracted from the test corpus. These sequences encompass diverse recovery scenarios including lie-to-stand, prone-to-stand, and stand-to-lie transitions, with varied initial fallen configurations and recovery directions. All sequences are evaluated in both their original continuous form and discretized variants, in which the trajectories are replaced with piecewise-constant poses, to assess robustness under discontinuous reference signals. Quantitative results are summarized in \cref{tab:getup}; qualitative sim-to-sim comparisons are shown in \cref{fig:lie}.

\begin{table*}[!htbp]
\centering
\captionsetup{justification=raggedright,singlelinecheck=false}
\caption{\textbf{Fall-and-recovery evaluation on challenging lie-to-stand, prone-to-stand, and stand-to-lie sequences.} CR denotes completion rate. \colorbox{bestblue}{Blue} marks the best result per metric. Colored percentages show relative change from MLP: \textcolor{teal!80!black}{teal}\,=\,better, \textcolor{red!70!black}{red}\,=\,worse.}
\label{tab:getup}
\vspace{-2pt}
\footnotesize
\begin{tabular}{@{}lccccc@{}}
\toprule
\textbf{Method}
  & \makecell{\textbf{CR}\\(\%) $\uparrow$}
  & \makecell{\textbf{Joint Err}\\(rad) $\downarrow$}
  & \makecell{\textbf{Height Err}\\(m) $\downarrow$}
  & \makecell{\textbf{Ori Err}\\(rad) $\downarrow$}
  & \makecell{\textbf{LinVel Err}\\(m/s) $\downarrow$} \\
\midrule
MLP              & $44.0$ & $2.1720$ & $0.3586$ & $1.0157$ & $0.2710$ \\
\rowcolor{altrow}
Transformer      & $40.6$\,{\color{red}\scriptsize($-7.7\%$)} & $2.8309$\,{\color{red}\scriptsize($-30.3\%$)} & $0.3706$\,{\color{red}\scriptsize($-3.3\%$)} & $0.8355$\,{\color{teal}\scriptsize($+17.7\%$)} & $0.2708$\,{\color{teal}\scriptsize($+0.1\%$)} \\
VQ-VAE           & $69.8$\,{\color{teal}\scriptsize($+58.6\%$)} & $2.5700$\,{\color{red}\scriptsize($-18.3\%$)} & $0.1898$\,{\color{teal}\scriptsize($+47.1\%$)} & $0.4307$\,{\color{teal}\scriptsize($+57.6\%$)} & $0.2719$\,{\color{red}\scriptsize($-0.3\%$)} \\
\rowcolor{altrow}
SONIC            & $42.8$\,{\color{red}\scriptsize($-2.7\%$)} & $2.9897$\,{\color{red}\scriptsize($-37.6\%$)} & $0.3342$\,{\color{teal}\scriptsize($+6.8\%$)} & $0.8629$\,{\color{teal}\scriptsize($+15.0\%$)} & $0.2895$\,{\color{red}\scriptsize($-6.8\%$)} \\
\addlinespace[3pt]
iFSQ$_{\text{BM}}$  & $52.4$\,{\color{teal}\scriptsize($+19.1\%$)} & $2.4482$\,{\color{red}\scriptsize($-12.7\%$)} & $0.2880$\,{\color{teal}\scriptsize($+19.7\%$)} & $0.9109$\,{\color{teal}\scriptsize($+10.3\%$)} & $0.3134$\,{\color{red}\scriptsize($-15.6\%$)} \\
\rowcolor{altrow}
iFSQ$_{\text{+H}}$  & $52.7$\,{\color{teal}\scriptsize($+19.8\%$)} & $1.7660$\,{\color{teal}\scriptsize($+18.7\%$)} & \cellcolor{bestblue}$\mathbf{0.0419}$\,{\color{teal}\scriptsize($+88.3\%$)} & $0.2744$\,{\color{teal}\scriptsize($+73.0\%$)} & $0.2793$\,{\color{red}\scriptsize($-3.1\%$)} \\
iFSQ             & $48.2$\,{\color{teal}\scriptsize($+9.5\%$)} & $2.0236$\,{\color{teal}\scriptsize($+6.8\%$)} & $0.3024$\,{\color{teal}\scriptsize($+15.7\%$)} & $1.0488$\,{\color{red}\scriptsize($-3.3\%$)} & \cellcolor{bestblue}$\mathbf{0.2405}$\,{\color{teal}\scriptsize($+11.3\%$)} \\
\rowcolor{altrow}
Heracles         & \cellcolor{bestblue}$\mathbf{90.0}$\,{\color{teal}\scriptsize($+104.5\%$)} & \cellcolor{bestblue}$\mathbf{1.4114}$\,{\color{teal}\scriptsize($+35.0\%$)} & $0.0762$\,{\color{teal}\scriptsize($+78.7\%$)} & \cellcolor{bestblue}$\mathbf{0.2427}$\,{\color{teal}\scriptsize($+76.1\%$)} & $0.2830$\,{\color{red}\scriptsize($-4.4\%$)} \\
\bottomrule
\end{tabular}
\end{table*}

The fall-and-recovery evaluation reveals a stark performance divide that underscores the fundamental limitations of pure tracking paradigms (\cref{tab:getup}). All reactive tracking baselines---MLP, Transformer, and SONIC---achieve completion rates below 45\%, indicating complete inability to execute fall-recovery motions. While these methods function adequately under nominal tracking conditions (\cref{tab:comparison}), they fail catastrophically when the reference motion demands transitions through extreme pose configurations inherent to fall-and-recovery sequences.

Among the compared methods, VQ-VAE demonstrates unexpected resilience (CR\,=\,69.8\%), suggesting that its less precise but more flexible latent representation provides some implicit generalization to extreme poses. The iFSQ tracker variants without the generative middleware show mixed results: iFSQ$_{\text{+H}}$ achieves the highest completion rate among standalone trackers (52.7\%) and the lowest height error (0.0419\,m) due to its explicit height features, while iFSQ$_{\text{BM}}$ and iFSQ achieve completion rates of 52.4\% and 48.2\% respectively with substantially higher tracking errors.

\textbf{Heracles achieves the highest completion rate by a decisive margin} (90.0\%), exceeding the second-best method VQ-VAE by 20.2 percentage points---a 104.5\% relative improvement over MLP. Critically, Heracles also attains the lowest joint-position error (1.4114\,rad) and orientation error (0.2427\,rad) among all methods, confirming that the state-conditioned generative middleware is essential for maintaining coherent tracking through the extreme state transitions characteristic of fall-and-recovery motions. By dynamically synthesizing feasible recovery trajectories conditioned on the robot's real-time physical state, Heracles bridges the gap between the reference motion and the robot's actual configuration, enabling graceful execution of motions that drive purely reactive trackers to catastrophic failure.

\FloatBarrier
\subsection{Ablation Studies and Architectural Analysis}

To isolate the contribution of each design choice within the generative middleware, we conduct ablation experiments on the trajectory generator while keeping the iFSQ tracker fixed. All ablations are evaluated on the full 101-sequence test set spanning the complete diversity of the evaluation corpus. Results are summarized in \cref{tab:ablation}.

\begin{table*}[!htbp]
\centering
\caption{\textbf{Ablation study on the trajectory generator's key design components.} All variants are evaluated on the full 101-sequence test set using the same iFSQ tracker. \colorbox{bestblue}{Blue} marks the best result per metric. Colored percentages show relative change from Heracles (full): \textcolor{teal!80!black}{teal}\,=\,better, \textcolor{red!70!black}{red}\,=\,worse.}
\label{tab:ablation}
\vspace{-2pt}
\footnotesize
\begin{tabular}{@{}lccccc@{}}
\toprule
\textbf{Variant}
  & \makecell{\textbf{CR}\\(\%) $\uparrow$}
  & \makecell{\textbf{Joint Err}\\(rad) $\downarrow$}
  & \makecell{\textbf{Height Err}\\(m) $\downarrow$}
  & \makecell{\textbf{Ori Err}\\(rad) $\downarrow$}
  & \makecell{\textbf{LinVel Err}\\(m/s) $\downarrow$} \\
\midrule
Heracles (full)                 & \cellcolor{bestblue}$\mathbf{90.6}$ & \cellcolor{bestblue}$\mathbf{1.3272}$ & \cellcolor{bestblue}$\mathbf{0.0764}$ & \cellcolor{bestblue}$\mathbf{0.2728}$ & $0.2325$ \\
\addlinespace[3pt]
w/o directional warm start      & $87.2$ \pctb{-3.8} & $1.6236$ \pctb{-22.3} & $0.0962$ \pctb{-25.9} & $0.3393$ \pctb{-24.4} & $0.2423$ \pctb{-4.2} \\
\rowcolor{altrow}
w/o noisy-state augmentation    & $78.6$ \pctb{-13.2} & $1.8896$ \pctb{-42.4} & $0.1463$ \pctb{-91.5} & $0.4318$ \pctb{-58.3} & \cellcolor{bestblue}$\mathbf{0.2182}$ \pctg{+6.1} \\
w/o kinematics-aware weighting  & $82.1$ \pctb{-9.4} & $1.6931$ \pctb{-27.6} & $0.1200$ \pctb{-57.1} & $0.4055$ \pctb{-48.6} & $0.2394$ \pctb{-3.0} \\
\bottomrule
\end{tabular}
\end{table*}

\textbf{Directional Warm Start.}
Replacing the directional motion prior (\cref{eq:method_sdedit}) with pure Gaussian initialization degrades all metrics: completion drops from 90.6\% to 87.2\% ($-$3.8\%) and joint error increases by 22.3\%. The warm start seeds the ODE solver with a coarse linear interpolation toward the target, enabling the learned velocity field to focus its refinement budget on naturalness rather than gross direction estimation. Without this prior, the generator must expend additional integration steps to discover the correct recovery heading, yielding failures particularly on fall-and-recovery sequences.

\textbf{Noisy-State Augmentation.}
Removing the asymmetric noise injection (\cref{eq:method_noise}) during training produces the most severe degradation in completion rate among all ablation variants, with CR falling to 78.6\% ($-$13.2\%) and joint error increasing by 42.4\%. Height error nearly doubles ($+$91.5\%), and orientation error increases by 58.3\%. Without noise augmentation, the generator overfits to clean state inputs; at deployment, accumulated tracking drift and sensor noise push the conditioning state away from the training distribution, causing catastrophic failure on out-of-distribution motions. The augmented variant bridges this train--deploy distribution gap, enabling robust trajectory generation even from noisy proprioceptive readings.

\textbf{Kinematics-Aware Loss Weighting.}
Removing the Jacobian-based weighting (\cref{eq:method_jac}) produces the largest degradation in completion rate after noisy-state augmentation, with CR falling to 82.1\% ($-$9.4\%). Height error increases by 57.1\% and orientation error by 48.6\%. The severity of this ablation indicates that pose-dependent lever-arm geometry is critical for robust tracking: a unit-radian shoulder error in an extended-arm configuration induces far larger Cartesian displacement than the same error with arms at rest, and the weighting scheme enables the generator to prioritize these geometrically sensitive configurations.

\begin{figure}[!htbp]
\centering
\includegraphics[width=\columnwidth]{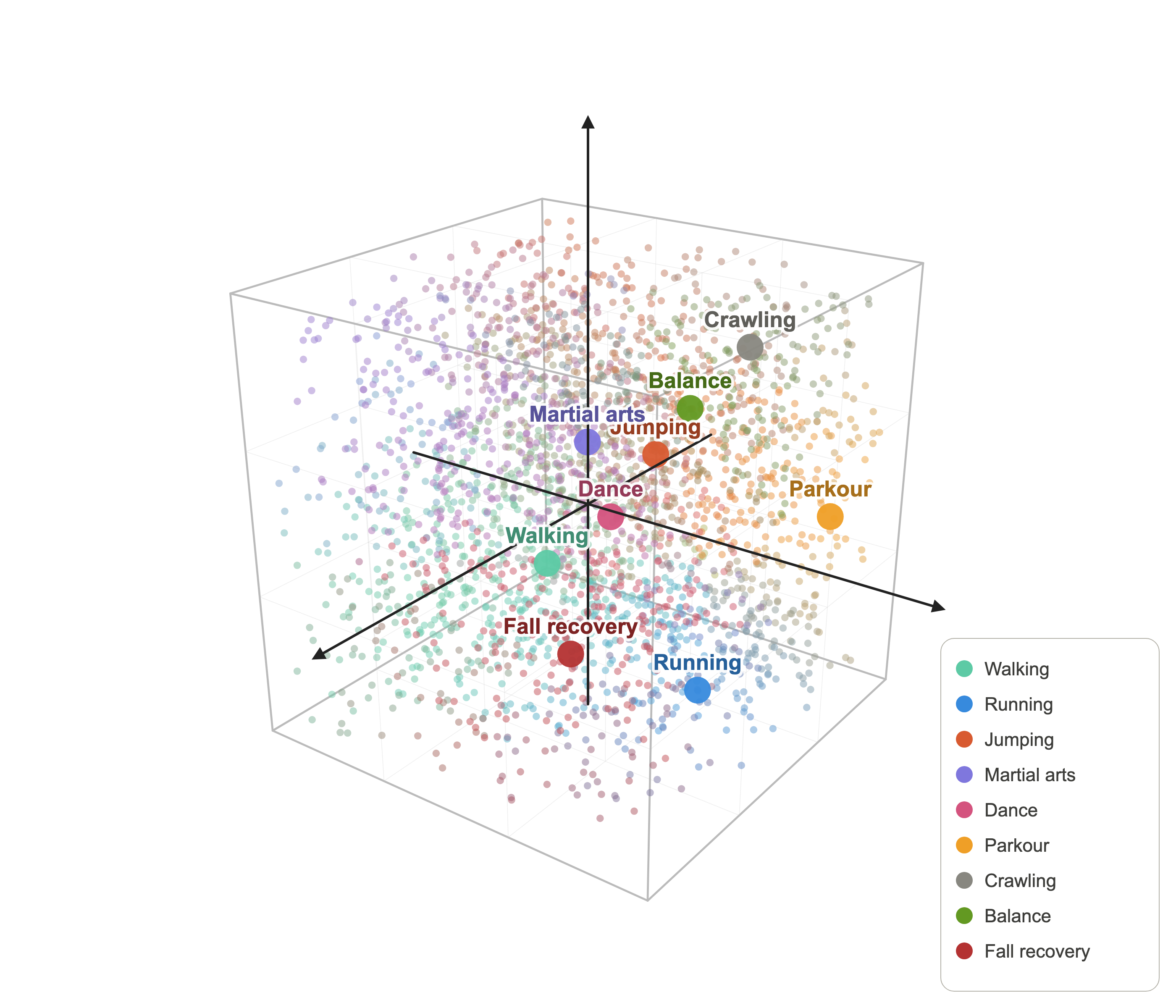}
\captionsetup{justification=raggedright,singlelinecheck=false}
\caption{\textbf{Emergent semantic clustering in the learned iFSQ codebook.} Each point represents a quantized motion token projected into 3D via PCA; colors denote motion categories. Despite receiving no category labels during training, the codebook self-organizes into semantically coherent clusters corresponding to distinct motor skills.}
\label{fig:fsq_codebook}
\end{figure}

\textbf{Learned Discrete Representation.}
Beyond component-level ablations, we examine whether the iFSQ tokenizer acquires a semantically structured codebook after training on the full motion corpus. \cref{fig:fsq_codebook} visualizes the discrete code activations projected into a three-dimensional embedding space, with each point representing a quantized motion token colored by its source motion category. The visualization reveals clearly separable clusters corresponding to distinct motor skills---walking, running, jumping, martial arts, dance, parkour, crawling, balance, and fall recovery---despite the quantizer receiving no explicit category labels during training. Notably, semantically related skills occupy neighboring regions (walking and running clusters lie adjacent, while crawling and balance form a separate group), indicating that the iFSQ codebook captures meaningful kinematic similarity structure. This emergent organization confirms that the finite scalar quantization not only compresses high-frequency motion signals into compact tokens but also distills a structured motion taxonomy that enables the downstream action decoder and trajectory generator to reason over semantically coherent motion abstractions.

\textbf{Architectural Motivation.}
The ablation results collectively reveal why a hierarchical architecture---with a dedicated generative planner layered above a physics tracker---is preferable to monolithic alternatives for general-purpose humanoid control. On the 101-sequence evaluation, removing any single component reduces completion rate by 3.8--13.2\%, demonstrating that all three design choices are essential for robust performance. Noisy-state augmentation has the largest impact on tracking quality (CR: $-$13.2\%, height error: $+$91.5\%), underscoring that bridging the train--deploy distribution gap is the most critical challenge for sustained tracking fidelity. Kinematics-aware weighting produces the second-largest CR degradation ($-$9.4\%) alongside substantial increases in height ($+$57.1\%) and orientation error ($+$48.6\%), revealing that accurate lever-arm modeling is essential for navigating complex multi-step transitions. These findings reinforce the design principle of frequency separation: the generative middleware reasons over a 0.2\,s planning window at 25\,Hz, synthesizing temporally coherent recovery strategies that the tracker then faithfully executes at 50\,Hz---mirroring the hierarchical structure of biological motor systems.

\FloatBarrier
\subsection{Recovery Behaviors and Analysis}


Beyond quantitative metrics, we examine the qualitative character of recovery behaviors to understand \emph{how} the generative middleware transforms the robot's response to severe disturbances. This analysis reveals a fundamental distinction between tracking-based and generation-based control paradigms.

\textbf{Tracking-Only Failure Modes.}
When subjected to large external perturbations, standalone trackers---including all baselines and the iFSQ variant---exhibit a characteristic failure pattern. Upon being displaced from the reference trajectory, the tracker computes a single-step corrective action that minimizes the instantaneous state-reference error. This myopic strategy produces rigid, jerky corrective torques that lack the temporal coordination required for dynamic balance recovery. In the most severe cases, the tracker's insistence on returning to the exact reference pose forces physically infeasible joint configurations, accelerating rather than preventing the fall. Even when the tracker avoids catastrophic failure, its recovery motions appear distinctly non-anthropomorphic: abrupt whole-body stiffening, unnatural arm postures, and a conspicuous absence of the compensatory stepping strategies that characterize human balance recovery.

\textbf{Generative Recovery Behaviors.}
Heracles produces qualitatively different recovery dynamics. When a large perturbation displaces the robot from the reference manifold, the generative middleware detects the state-reference discrepancy and synthesizes a short-horizon trajectory that prioritizes physical feasibility over immediate reference fidelity. This manifests as emergent human-like recovery strategies: compensatory stepping to widen the base of support, coordinated arm counter-motions to redistribute angular momentum, and gradual torso realignment before resuming the original motion. Crucially, these behaviors are not hand-designed or reward-engineered---they emerge naturally from the flow matching model's learned distribution over physically plausible motion transitions, conditioned on the robot's real-time state.

\textbf{From Tracking to Planning: Rethinking General Humanoid Control.}
The observed behavioral difference reveals a deeper conceptual insight into what constitutes a truly general-purpose humanoid controller. The dominant tracking paradigm implicitly assumes that control reduces to minimizing the deviation between the robot's state and a predefined kinematic reference. While effective for nominal execution, this formulation conflates two fundamentally distinct objectives: executing a desired motor intent and maintaining physical viability.

Human motor control does not operate as a rigid reference tracker. When a person stumbles, they do not attempt to snap back to a pre-planned gait trajectory. Instead, the motor system rapidly \emph{revises} the intended trajectory itself, generating a new plan that accounts for the current physical state, gravitational constraints, and available momentum. The original intent is temporarily deprioritized in favor of a dynamically feasible recovery path, and only once stability is restored does the system smoothly re-engage with the original task objective.

Heracles embodies precisely this principle through its state-conditioned middleware. The generative planner continuously modulates the reference signal based on real-time physical feasibility: passing commands through unmodified when tracking is viable, but seamlessly \emph{rewriting} them when the physical state demands a different motor strategy. This transforms the controller from a passive trajectory follower into an active trajectory synthesizer that reasons about what the robot \emph{should} do given its current physical reality, rather than what it was \emph{told} to do by a reference signal computed without knowledge of real-time dynamics.

\begin{figure}[!htbp]
\centering
\captionsetup{justification=raggedright,singlelinecheck=false}
\includegraphics[width=\columnwidth]{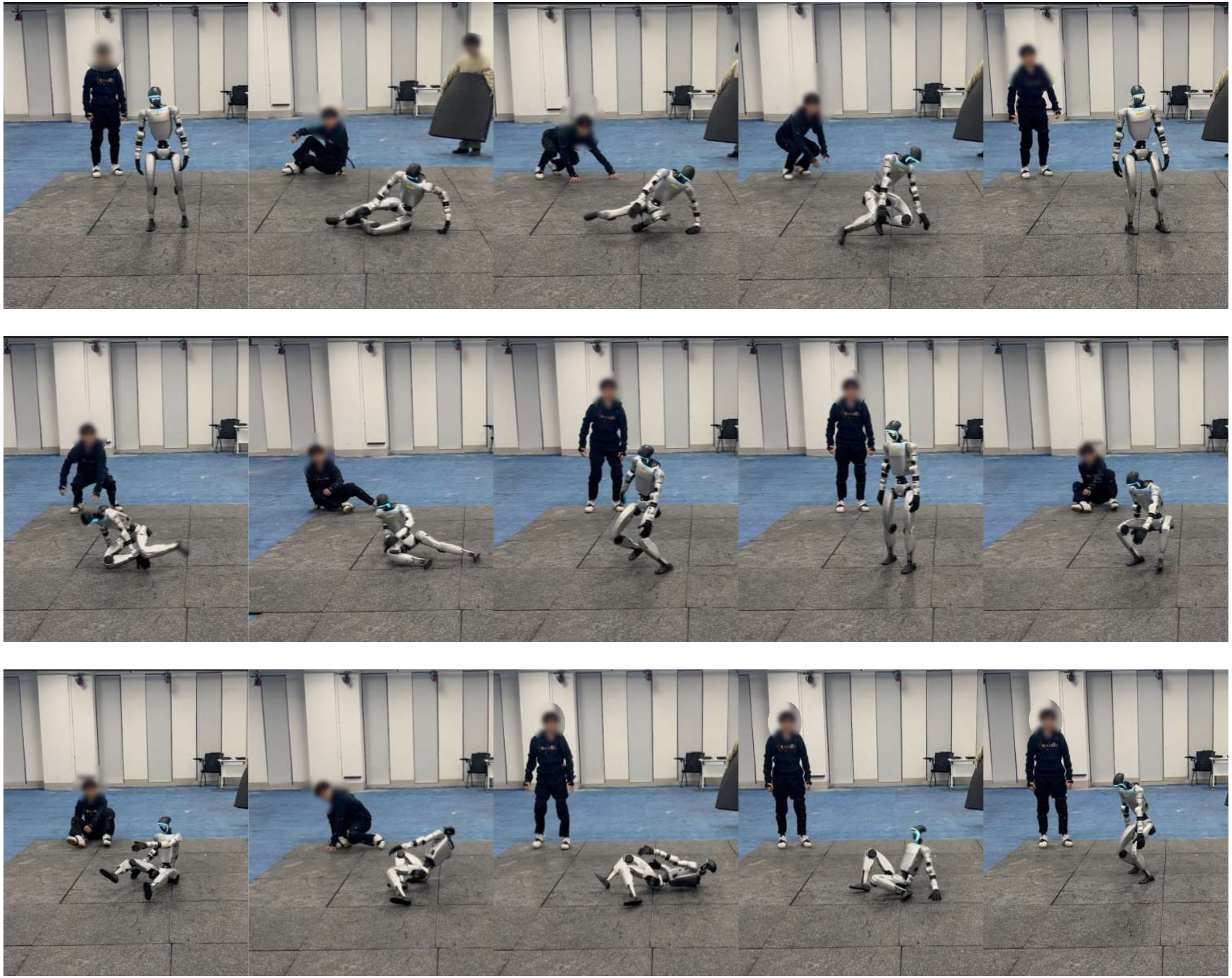}
\caption{\textbf{Omnidirectional fall recovery motion tracking.} Real-world experiments demonstrate that our model generalizes to arbitrary lie-to-stand recovery motions, successfully handling varied initial fallen configurations and recovery directions without task-specific engineering.}
\label{fig:real}
\end{figure}

\textbf{Implications for General-Purpose Deployment.}
This paradigm shift has concrete implications for deploying humanoid robots in unstructured environments. Real-world scenarios invariably introduce perturbations absent from any training distribution: unexpected collisions, terrain irregularities, payload changes, or degraded actuation. A tracking-only controller can only succeed if its training-time domain randomization happens to cover the encountered disturbance; beyond this envelope, it fails brittly. The generative middleware, by contrast, provides a principled mechanism for \emph{open-ended} adaptation: as long as the learned motion prior contains transitions between sufficiently diverse physical states, the model can compose novel recovery strategies for previously unseen perturbations. This compositional generalization---the ability to recombine learned motion primitives into new sequences conditioned on novel states---is what distinguishes a truly general controller from one that merely covers a large but finite set of pre-trained behaviors.

We further evaluate this generalization capability on omnidirectional fall recovery tasks in both simulation and the real world. As shown in \cref{fig:lie}, Heracles completes the full lie-to-stand transition in MuJoCo while all baseline methods fail or only partially succeed. \cref{fig:real} presents the corresponding real-world results: across three trials, the robot is initialized in distinct fallen configurations---supine, lateral, and prone postures. In all cases, the robot successfully executes a complete lie-to-stand recovery by following the reference motion, progressively transitioning through intermediate support phases. Critically, the recovery directions vary across trials, with the robot rising toward different orientations relative to its initial fallen heading, demonstrating true omnidirectional recovery rather than a memorized fixed-direction strategy.

%% file: sections/5_conclusion.tex
\section{Conclusion}
In conclusion, this work presents Heracles, a state-conditioned generative middleware that fundamentally resolves the longstanding dichotomy between strict kinematic tracking and robust physical recovery in humanoid robotics. By embedding a continuous flow matching process within a closed-loop control architecture, the framework dynamically bridges high-fidelity intent execution with anthropomorphic resilience. Without relying on explicit mode-switching heuristics, Heracles intrinsically preserves exact zero-shot tracking under nominal conditions while seamlessly synthesizing dynamically feasible recovery maneuvers during severe environmental perturbations. Ultimately, this unified paradigm liberates embodied control from rigid, reference-bound execution, establishing a highly scalable foundation for deploying agile and resilient general-purpose humanoids in complex physical environments.

%% file: sections/6_project_team.tex
\section{X-Humanoid Heracles Project Team}

This report reflects a collaborative effort by the X-Humanoid Heracles project team. The roles and contributors are listed below.

\noindent\textbf{Project Leader.} Qiang Zhang

\noindent\textbf{Equal Contribution.} Zelin Tao, Zeran Su

\noindent\textbf{Project Team Members.} Peiran Liu, Jingkai Sun, Wenqiang Que, Jiahao Ma, Jialin Yu, Jiahang Cao, Pihai Sun, Hao Liang

\noindent\textbf{Technical Support.} Gang Han, Wen Zhao, Zhiyuan Xu, Yijie Guo, Jian Tang